\documentclass[lettersize,journal]{IEEEtran}
\usepackage{amsmath,amsfonts}
\usepackage{algorithmic}
\usepackage{algorithm}
\usepackage{array}

\usepackage{amssymb,enumitem}

\usepackage{subcaption}     
\usepackage[table]{xcolor}	
\usepackage{textcomp}
\usepackage{stfloats}
\usepackage{url}
\usepackage{verbatim}
\usepackage{graphicx}
\usepackage{cite}
\usepackage{nccmath}
\usepackage[overload]{empheq}
\usepackage{MnSymbol}
\newcommand{\norm}[1]{\left\lVert#1\right\rVert}    

\newcommand{\R}{\mathbb{R}}                         
\newcommand{\C}{\mathbb{C}}                         

\usepackage{ntheorem}
{}
{}
\newtheorem{prop}{Proposition}{}
\newtheorem{prob}{Problem}{}
\newtheorem{thm}{Theorem}{}
{}
\newtheorem{lem}{Lemma}{}
\newtheorem{assum}{Assumption}{}

\begin{document}

\title{Hybrid Model-Data Fault Diagnosis for Wafer Handler Robots: Tilt and Broken Belt Cases}

\author{Tim van Esch$^{1}$, Farhad Ghanipoor$^{2}$, Carlos Murguia$^{2,3}$, and Nathan van de Wouw$^{2}$

\thanks{$*$  The first and second authors contributed equally to this work.}        
\thanks{$^{1}$ ASML, Veldhoven, the Netherlands. E-mail: t.vanesch@hotmail.com.}
\thanks{$^{2}$ Mechanical Engineering Department, Eindhoven University of Technology, Eindhoven, the Netherlands. E-mails: f.ghanipoor@tue.nl, c.g.murguia@tue.nl, and n.v.d.wouw@tue.nl.}
\thanks{$^{3}$ Engineering Systems Design, Singapore University of Technology and Design, Singapore. E-mail: {murguia\_rendon@sutd.edu.sg}.}}


\maketitle

\begin{abstract}
This work proposes a hybrid model- and data-based scheme for fault detection, isolation, and estimation (FDIE) for a class of wafer handler (WH) robots. The proposed hybrid scheme consists of: 1) a linear filter that simultaneously estimates system states and fault-induced signals from sensing and actuation data; and 2) a data-driven classifier, in the form of a support vector machine (SVM), that detects and isolates the fault type using estimates generated by the filter. We demonstrate the effectiveness of the scheme for two critical fault types for WH robots used in the semiconductor industry: broken-belt in the lower arm of the WH robot (an abrupt fault) and tilt in the robot arms (an incipient fault). We derive explicit models of the robot motion dynamics induced by these faults and test the diagnostics scheme in a realistic simulation-based case study. These case study results demonstrate that the proposed hybrid FDIE scheme achieves superior performance compared to purely data-driven methods.

\end{abstract}


\section{Introduction} \label{ch:introduction}
In the semiconductor industry, there is a broad interest in the application of advanced health monitoring (HM) algorithms for wafer handler (WH) robots used in lithography equipment. These high-end robots transfer wafers between different stations in the chip manufacturing process and are operated at a high speed while maintaining high positioning accuracy to maximize the productivity of these processes \cite{cong2007research,cong2009wafer,lee2023system}. Currently, unexpected downtime of WH robots requires unscheduled maintenance of the lithography machines, which is time-consuming and expensive as it stops the entire manufacturing process. To remedy these unexpected failures, fault detection, isolation, and estimation (FDIE) techniques have become increasingly important. With proper fault diagnosis, the reliability and cost-efficiency of these machines can be increased drastically. In addition, if the severity of faults could be estimated and used to predict when the robot would fail, then, it would become possible to change unscheduled downtime into (much affordable) scheduled downtime.

This paper studies the FDIE problem for two critical fault scenarios in WH robots: broken belt in the lower arm and tilt in the robot arms.
\color{black}Given physics-based (Euler-Lagrange) models for the motion dynamics of wafer handler robots, model-based methods for FDIE have been studied extensively in the literature for robots with similar system dynamics, but for different fault types \cite{lee2023system,capisani2010higher,kim2020phase}, and for general dynamical systems \cite{chen2012robust,ding2008model,hwang2009survey}. In model-based FDIE, models of the first-principles physics of the system are used to monitor system health. Physics-based models are used to predict the system's expected behavior in response to driving inputs and initial conditions. These predictions are compared to real system data (obtained through sensor measurements) to generate the so-called monitoring residuals (the difference between measurements and model-based predictions). Residuals are subsequently used to detect mismatches potentially indicating the presence and location of faults in the system \cite{chen2012robust,ding2008model,hwang2009survey}. 

Faults can affect different entries (i.e., different equations) of system models. A known limitation of model-based methods is their inability to distinguish different types of faults that influence the same entry in the dynamics \cite{cheng2015combined,van2022multiple,ghanipoor2023robust}. This limitation arises because model-based methods typically isolate faults by either preserving or amplifying the effects of signals associated with entry of one fault, while minimizing or nullifying the effects of signals related to the entry of another fault on a specific residual. Clearly, if two faults share the same entry into the dynamics, it is impossible to both minimize and maximize it simultaneously. Unfortunately, the fault scenarios considered in this work (i.e., tilt and broken belt for WH robots) share the same entry in the WH dynamics (thereby making off-the-shelf model-based methods overly not applicable for fault isolation \cite{cheng2015combined, dong2023robust}). 

As opposed to model-based FDIE, data-driven FDIE methods perform fault diagnosis solely based on data. The strengths of data-driven methods lie within their capability of feature extraction from system measurements without model knowledge of the underlying system dynamics \cite{yin2014review}. Feature extraction makes it possible to identify certain faults based on system data through a range of ML algorithms. Data-driven methods have shown accurate diagnostics results in many applications \cite{lei2020applications}.
Unlike model-based techniques, data-based methods can handle faults with the same entry, as they rely only on fault features. As long as each fault has a unique feature in the available signals, these methods can classify them correctly \cite{van2022multiple}. Therefore, data-based methods are more suitable for the considered fault scenarios within WH robots. \color{black}

More recently, the interest of combining model-based and data-driven FDIE methods has been increasing. These so-called hybrid FDIE approaches aim to leverage the benefits of both model-based and data-driven FDIE techniques in order to devise novel high-performance FDIE schemes \cite{khorasgani2018framework,tidriri2016bridging,jung2016combined}. To arrive at a hybrid FDIE scheme, several components of model-based and data-driven approaches can be combined in different ways. State-of-the-art hybrid FDIE schemes combine a model-based approach for residual generation and a data-driven method for feature extraction to diagnose faults \cite{khorasgani2018framework}. 
The information of both algorithms can be supplied to a classifier which will then perform the decision-making process (information fusion) \cite{sheibat2014support}. Alternatively, a model-based and data-driven FDIE algorithm could be run in parallel. Both algorithms are then used to perform the FDIE decision-making process; these decisions will be compared to each other (decision fusion) to arrive at a final decision \cite{tidriri2018generic}. 

Given the fact that model-based methods cannot isolate the fault scenarios considered here for WH robots, we propose a novel hybrid scheme that combines a model-based fault estimator with a data-based method to enable fault isolation (FI). The proposed fault estimator offers additional benefits compared to common residual generators, as it estimates the fault signal entering the system dynamics. Moreover, we compare the proposed method to a purely data-based approach to demonstrate its superior performance in the application to an industrial WH robot. This work is evaluated in a simulation-based case study, as introducing the studied faults on the actual robot would require destructive tests that should be avoided at this stage.
\color{black}

	The main contributions of this paper are as follows: 
	\begin{enumerate} [label=(\alph*)]
		\item 	 \textbf{\emph{Modeling of Critical Faults in Wafer Handler Robots}}: Two critical faults in wafer handler robots, namely tilt in the robot arms and a broken belt in the lower arm, are modeled in terms of their impact on the robot dynamics.
        \item \textbf{\emph{Optimization-Based Fault Estimation Scheme}}: Our approach proposes a computationally efficient algorithm for synthesizing the design parameters of the proposed fault estimator. This is accomplished by solving semi-definite programs to minimize the $H_{\infty}$-norm of perturbations from (fault and nonlinearity) model mismatches to the fault estimation error. Furthermore, we provide tractable conditions for stability guarantees while imposing desired upper bounds on the $H_2$-norm from measurement noise to the fault estimation error (Theorem~\ref{theorem:optimal_estimator}). These tractable design conditions allow for performance trade-off analyses, enhancing robustness against various perturbations.
		\item \textbf{\emph{Isolation of Same-Entry Faults}}: The proposed hybrid method for fault isolation overcomes the limitations of conventional model-based methods by isolating faults that affect the same system entries \cite{cheng2015combined, dong2023robust}. This is achieved by using data-driven classifiers for fault isolation, which can differentiate between various behaviors in the fault estimates produced by the model-based fault estimator. This capability is crucial for isolating tilt and broken belt faults, which impact the same entry in wafer handler robot dynamics.
        \color{black}
    	\item \textbf{\emph{Superior Performance Compared to Purely Data-Driven Methods}}: We show that the proposed hybrid model-data fault diagnosis scheme for the wafer handler robots outperforms purely data-driven methods in fault isolation tasks by exploiting fault estimates, generated by the above filter, instead of direct raw input-output data for fault isolation.
	\end{enumerate}
 

The estimation algorithms presented in the paper build the results in \cite{ghanipoor2023linear}, where the authors present a fault estimation scheme for nonlinear dynamical systems that leverages robust control tools and ultra-local modeling techniques.  Compared to \cite{ghanipoor2023linear}, this paper provides all the technical proofs of the technical results regarding the estimator design, not included in \cite{ghanipoor2023linear}. In addition, this paper contributes on a data-based method for fault isolation, exploiting the fault estimates. By integrating the model-based fault estimation method, a novel hybrid FDIE scheme is proposed. Moreover, the proposed method is evaluated on a an industrial use case of wafer handler robots, exhibiting highly nonlinear dynamics and two critical, multiplicative fault types.
\color{black}

The outline of the paper is as follows. Section~\ref{ch:modeling} introduces the considered dynamics of the WH robot and the considered failure modes. Section~\ref{ch:estimation} covers the design of a model-based fault estimator for the WH robot. Simulation-based results are provided for the designed fault estimator. Section~\ref{ch:hybrid} presents fault detection and isolation results using machine-learning-based classification methods. These techniques are integrated with the model-based fault estimator, as developed in Section~\ref{ch:estimation}, into a hybrid FDIE scheme. Then, the proposed hybrid FDIE scheme is evaluated in a simulation-based use case for the wafer handler robot to address a wide range of fault scenarios. Additionally, the performance of the proposed hybrid approach is compared to a purely data-driven approach for fault isolation. Section~\ref{ch:conclusion} contains the conclusions and recommendations for future work.

\textbf{Notation:} The $n \times n$ identity matrix is denoted by $I_n$ or simply $I$ if $n$ is clear from the context. Similarly, $n \times m$ matrices composed of only zeros are denoted by ${0}_{n \times m}$ or simply ${0}$. First and second time-derivatives of a vector $x$ are expressed as $\dot{x}$ and $\ddot{x}$, respectively. For higher-order time-derivatives of a vector $x$, the notation $x^{(r)}$ is adopted. A positive definite matrix is denoted by $X \succ 0$ and a positive semi-definite matrix is presented by $X \succeq 0$. Similarly, for a negative definite matrix, $X \prec 0$ is used, and $X \preceq 0$ for a negative semi-definite matrix. The imaginary unit $j$ is defined by $j^2 = -1$. For a transfer function $T(s)$, with $s \in \C$, $\sigma_{\max}\big(T(s) \big)$ denotes the maximum singular value, and $T^H (s)$ represents the Hermitian transpose. For a vector $x$, $\norm{x}$ represents the Euclidean norm. The notation $(x_1, \cdots, x_n)$ stands for the column vector composed of the scalar/vector elements $x_1,\ldots,x_n$. Finally, $x \sim \mathcal{U}(a,b)$ represents a vector $x$ whose elements are drawn from a uniform distribution bounded by the domain $[a,b]$. Often time-dependencies are omitted for notational simplicity.

\section{Modeling of the Robot Dynamics}\label{ch:modeling}
\begin{figure}[tbp]
    \centering
    \includegraphics[width=0.65\linewidth]{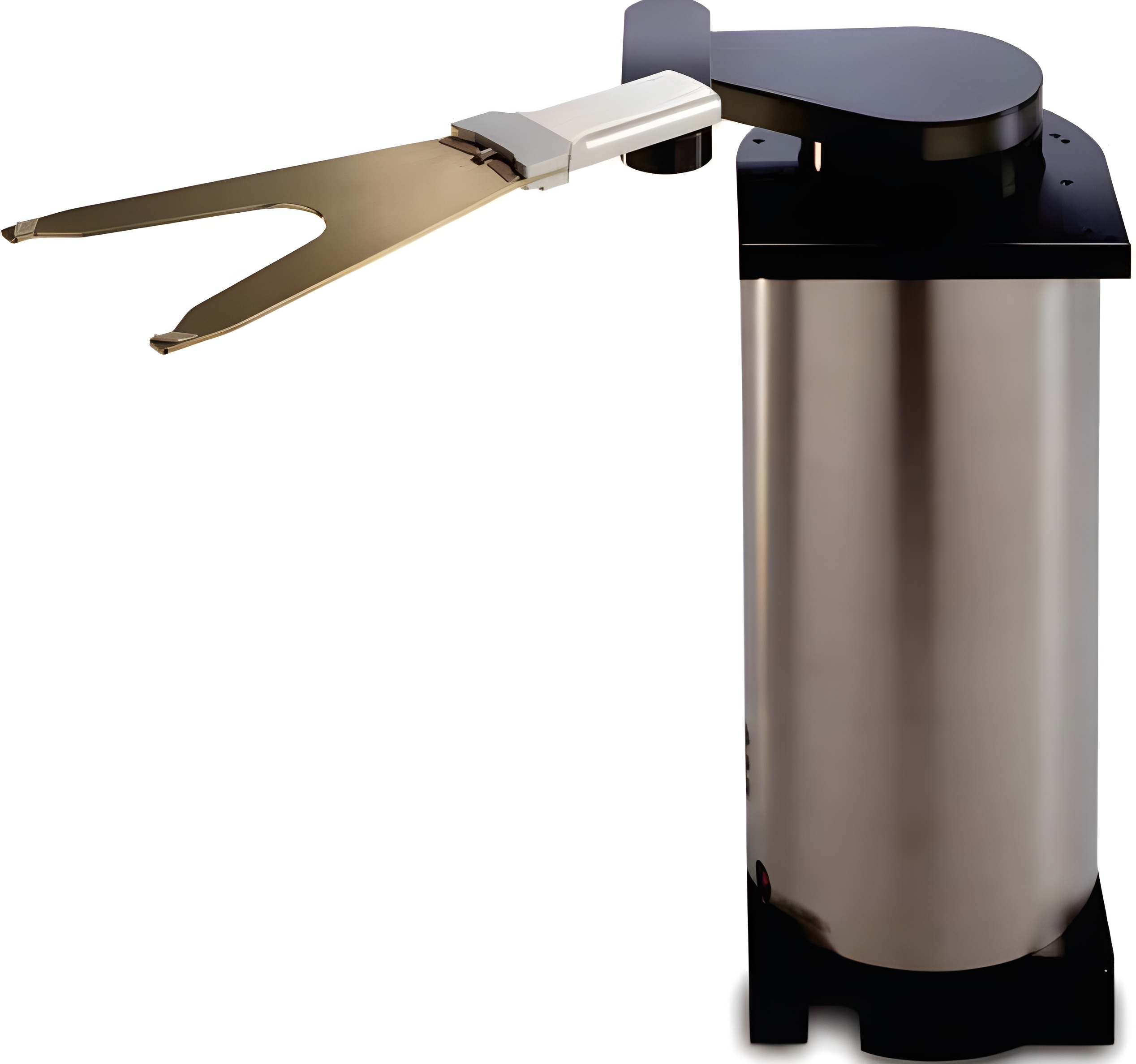}
    \caption{An example of SCARA manipulator \cite{zhang2007calibration}.}
    \label{fig:scara_robot}
\end{figure}

\begin{figure}[tbp]
    \centering
    \includegraphics[width=0.8\linewidth]{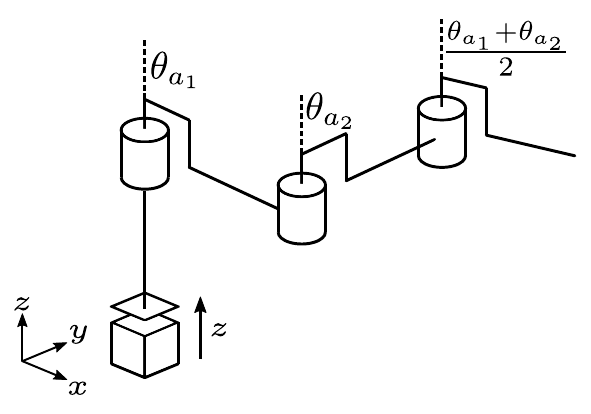}
    \caption{Schematic representation of a SCARA manipulator.}
    \label{fig:scara_schematic}
\end{figure}

In this section, the healthy dynamics (i.e., without faults) of the considered class of WH manipulators is characterized. The considered WH robot lies in the class of SCARA manipulators with cylindrical workspace. A picture of such a manipulator is given in Figure~\ref{fig:scara_robot}, and a schematic representation is depicted in Figure~\ref{fig:scara_schematic}. The angular position of the upper and lower arm is given by $\theta_{a_1}$ and $\theta_{a_2}$, respectively, and the angular position of the end-effector is determined by $\theta_{a_3} = (\theta_{a_1}+\theta_{a_2})/2$, due to a physical (holonomic) constraint between the links. Finally, the vertical position is denoted by $z$. The motors drive rotational degrees of freedom of the arms through a power transmission system and the angular positions of the motors are measured via encoders, which are co-located at the motors. Hence, the arm angles are not directly measured, but only indirectly via the dynamics of the motor and the connection to the arm. The dynamics of the WH robot can be split into two parts: the (nonlinear) planar dynamics and the linear vertical dynamics of the robot. Only the planar dynamics are considered here as these are deemed most useful for the fault diagnosis scenarios that are considered.

\subsection{Planar Dynamics}
\begin{figure}[tbp]
    \centering
    \includegraphics[width=\linewidth]{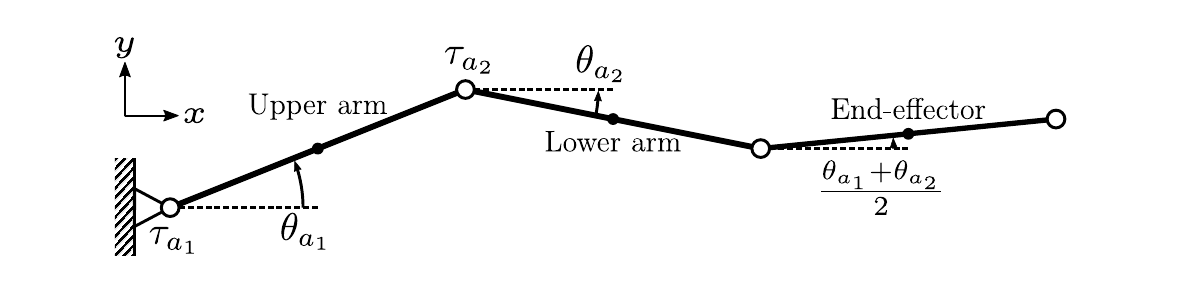}
    \caption{Planar schematic overview of the WH robot in the XY-plane.}
    \label{fig:arm_kinematics}
\end{figure}

The planar dynamics describe the radial movement of the WH manipulator. Due to this radial movement, the inertia of the system is dependent on the position of the links, resulting in nonlinear dynamics. A schematic overview of the planar manipulator is depicted in Figure~\ref{fig:arm_kinematics}. Considering the rigid planar arm dynamics, the equations of motion (EOM) of the WH manipulator (obtained following the Euler Lagrange formalism \cite{EulerLagrange}) can be written as
\begin{equation}
    \tau_a = \underline{M}(\theta_a)\ddot{\theta}_a + \underline{C}(\theta_a, \dot{\theta}_a),
    \label{eq:arm_kinematics}
\end{equation}
where $\theta_a := [\theta_{a_1} \ \theta_{a_2} ]^\top \in \mathbb{R}^2$ contains the link angular positions of the robot, $\dot{\theta}_a \in \mathbb{R}^2$ and $\ddot{\theta}_a \in \mathbb{R}^2$ denote angular velocities and accelerations, respectively, and $\tau_a := [ \tau_{a_1} \ \tau_{a_2} ]^\top \in \mathbb{R}^2$ are the torques acting on the joints of the arms. Moreover, $\underline{M}(\theta_a)\in \mathbb{R}^{2\times 2}$ denotes the position-dependent inertia matrix and $\underline{C}(\theta_a, \dot{\theta}_a) \in \mathbb{R}^{2}$ is the nonlinear vector of centrifugal and Coriolis forces. The full expressions of $\underline{M}(\theta_a)$ and $\underline{C}(\theta_a, \dot{\theta}_a)$ are given in Appendix~\ref{app:eom}. The torque supplied to the arm (i.e., $\tau_a$) represents the motor torque delivered via a power transmission. In what follows, the power transmission dynamics are described. 

\subsection{Power Transmission Dynamics}
\begin{figure}[tbp]
    \centering
    \includegraphics[width=0.8\linewidth]{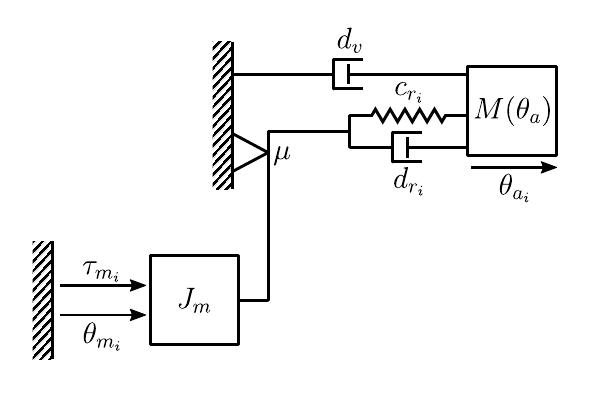}
    \caption{Schematic overview of the power transmission model.}
    \label{fig:harmonic_drive}
\end{figure}

A schematic overview of the power transmission dynamics is given in Figure~\ref{fig:harmonic_drive}. Including the power transmission dynamics into the model requires considering the rotational displacement of the motors $\theta_m := [\theta_{m_1} \ \theta_{m_2} ]^\top \in \mathbb{R}^2$. The dynamics of power transmission are derived in Appendix~\ref{app:gearbox_daynamics} and can be described by the following linear dynamics:
\begin{equation}
    \tau_m = \underline{J}\ddot{\theta}_m + \underline{D}(\mu^2 \dot{\theta}_m - \mu \dot{\theta}_a) + \underline{K}(\mu^2 \theta_m - \mu \theta_a),
    \label{eq:harmonic_drive}
\end{equation}
where the input torque from the motors is denoted as $\tau_m := [ \tau_{m_1} \ \tau_{m_2} ]^\top \in \mathbb{R}^2$, $\underline{J} \in \mathbb{R}^{2\times 2}$ is the diagonal inertia matrix containing the inertia of the motors $J_m$ on its diagonal, and $\mu \in \mathbb{R}$ is the transmission ratio of the gearbox in the power transmission. Matrix $\underline{D} \in \R^{2\times 2}$ denotes the diagonal damping matrix containing the damping coefficients $d_{r_i}$, and $\underline{K}\in \R^{2\times 2}$ is the diagonal stiffness matrix containing the stiffness coefficients $c_{r_i}$, with $i\in\{1,2\}$ indicating the corresponding link.

Furthermore, considering the power transmission schematic in Figure~\ref{fig:harmonic_drive}, the torque acting on the joints $\tau_a$ can also be expressed in terms of the motor and link coordinates as follows:
\begin{equation}
    \tau_a = \underline{D}(\mu \dot{\theta}_m - \dot{\theta}_a) + \underline{K}(\mu \theta_m - \theta_a) - \underline{D}_v\dot{\theta}_a,
    \label{eq:torque}
\end{equation}
where $\underline{D}_v$ is a diagonal damping matrix that contains the viscous friction terms $d_v$ on the diagonal. Combining the rigid model in \eqref{eq:arm_kinematics} with the expression for $\tau_a$ in \eqref{eq:torque} yields:
\begin{equation}
    \begin{split}
        \underline{M}(\theta_a)\ddot{\theta}_a + \underline{C}(\theta_a, \dot{\theta}_a) &+ \underline{D}_v\dot{\theta}_a =\\ &\underline{D}( \mu \dot{\theta}_m - \dot{\theta}_a) 
        + \underline{K}(\mu \theta_m - \theta_a).
    \end{split}
    \label{eq:arm_kinematics_flexible}
\end{equation}

\subsection{Total Planar Dynamics}
To obtain a complete model of the WH manipulator, the planar arm dynamics and the power transmission dynamics need to be integrated. The complete model is constructed by combining \eqref{eq:harmonic_drive} and \eqref{eq:arm_kinematics_flexible} into
\begin{equation}
\left\{
    \begin{split}
        &\underline{J}\ddot{\theta}_m + \underline{D}(\mu^2 \dot{\theta}_m - \mu \dot{\theta}_a) + \underline{K}(\mu^2 \theta_m - \mu \theta_a) = \tau_m, \\
        &\underline{M}(\theta_a)\ddot{\theta}_a + \underline{C}(\theta_a, \dot{\theta}_a) + \underline{D}_v\dot{\theta}_a =\\ &\hspace{35mm}\underline{D}( \mu \dot{\theta}_m - \dot{\theta}_a) 
        + \underline{K}(\mu \theta_m - \theta_a). \\
    \end{split}
\right.
    \label{eq:dynamics_complete}
\end{equation}
Now, the equations in \eqref{eq:dynamics_complete} describe the complete dynamics from the input torque $\tau_m$ to the angular positions of the motors $\theta_m$ and links $\theta_a$. 

\subsection{State-Space Representation of the Healthy Dynamics}

In this section, the second-order dynamics introduced in \eqref{eq:dynamics_complete} is reformulated as a first-order state-space system (this reformulation will be instrumental for the fault estimation algorithms presented in the upcoming sections). Define the state vector $x = \big[ x_1^\top \ \ x_2^\top \ \ x_3^\top \ \ x_4^\top \big]^\top := \big[ \theta_m^\top \ \ \theta_a^\top \ \ \dot{\theta}_m^\top \ \ \dot{\theta}_a^\top \big]^\top \in \R^{8}$. Hence, directly from \eqref{eq:dynamics_complete}, we can write the state space model as follows:
\begin{subequations}
\begin{equation}
    \dot{x} = f_h(x,\tau_m),
    \label{eq:dynamics_state_space}
\end{equation}
where 
\begin{equation} \label{eq:dynamics_state_space2}
	f_h(x,\tau_m) =
	\begin{bmatrix}
		x_3 \\
		x_4 \\
		\begin{aligned}
			&-\underline{J}^{-1} \big( 
			\underline{D}(\mu^2 x_3 - \mu x_4) \\
			&\quad+ \underline{K}(\mu^2 x_1 - \mu x_2) 
			\big) 
			+ \underline{J}^{-1} \tau_m
		\end{aligned} \\
		\begin{aligned}
			&-\underline{M}(x_2)^{-1} \big( 
			\underline{C}(x_2, x_4) 
			+ \underline{D}(x_4 - \mu x_3) \\
			&\quad+ \underline{K}(x_2 - \mu x_1) 
			+ \underline{D}_v x_4 
			\big)
		\end{aligned}
	\end{bmatrix},
\end{equation}
and the output equation (characterizing measurable states):
\begin{equation}
    y = Cx,
\end{equation}
where $y$ is comprised of the motor angles $\theta_{m_1}$ and $\theta_{m_2}$ (as the encoders are co-located), and thus $C$ is given by
\begin{equation*}
    C :=
    \begin{bmatrix}
        I_2 & {0}
    \end{bmatrix}.
\end{equation*}
\end{subequations}


\subsection{Fault Scenarios and Associated Fault Models}
We consider two practically relevant fault scenarios encountered in WH robots. The first scenario involves a broken belt in the lower arm of WH robots. These belts form the connection between the motors and joints and are tensioned to ensure a rigid connection. When a belt breaks, erroneous displacements are introduced in the horizontal plane of the robot, and the physical, holonomic constraint for the end-effector, which imposes $\theta_{a_3} = (\theta_{a_1} + \theta_{a_2})/2$, no longer holds. As a result, we model the faulty behavior of the WH robot in the broken belt scenario as a manipulator with an additional degree of freedom ($\theta_{a_3}$) compared to the healthy model of the robot. It is important to note that the end-effector in this faulty scenario is not affected by any external forces, and its movements are due to initial conditions and the coupling of its dynamics with other links. Let the state vector for this faulty dynamics be denoted by $x_f = \big[ x_1^\top \ \ x_2^\top \ \ x_3^\top \ \ x_4^\top \ x_{f_1} \ \ x_{f_2} \big]^\top :=\big[  \theta_m^\top \ \ \theta_a^\top  \ \ \dot{\theta}_m^\top \ \ \dot{\theta}_a^\top \ \ \theta_{a_3} \ \ \dot{\theta}_{a_3} \big]^\top \in \R^{10}$. The faulty dynamics motion model for the broken belt scenario can be derived using the Euler Lagrange equations \cite{craig2006introduction}, and is given by
\begin{subequations} \label{eq:faulty_dynamics_belt}
\begin{equation}    \label{eq:dynamics_state_space_belt}
    \dot{x}_f  =
    \begin{bmatrix}
    	\dot{x} \\ \dot{x}_{f_1} \\ \dot{x}_{f_2}
    \end{bmatrix}
= f_{f_1}(x_f,\tau_m),
\end{equation}
where 
\begin{equation} \label{eq:dynamics_state_space_belt2}
	f_{f_1}(x_f,\tau_m) = 
		\begin{bmatrix}
        f_h(x,\tau_m) + \begin{bmatrix} 0 & 0 & 0 & I_2\end{bmatrix}^\top f_1(x_f) \\
        x_{f_2} \\
				\begin{aligned}
						&- \begin{bmatrix} 0 & 0 & 1 \end{bmatrix} \underline{M}_{f_1}(x_2,x_{f_1})^{-1}\\
						& \big( \underline{C}_{f_1}(x_2,x_4,x_{f_1},x_{f_2}) \\
						& \quad + \underline{D}_{f_1}(x_3,x_4) \\ 
						& \quad + \underline{K}_{f_1}(x_1,x_2)  + {\underline{D}_v}_{f_1}(x_4) \big),
					\end{aligned} 
			\end{bmatrix},
\end{equation}
and $\underline{M}_{f_1}(x_2,x_{f_1})\in \R^{3\times 3}$ and $\underline{C}_{f_1}(x_2,x_4,x_{f_1}, x_{f_2}) \in \R^{3}$ denote the position-dependent inertia vector and the nonlinear Coriolis and centrifugal matrix of the faulty dynamics for the broken belt, respectively. 
Additionally, $\underline{D}_{f_1}(x_3, x_4) \in \R^{3 \times 3}$,  $\underline{K}_{f_1}(x_1, x_2) \in \R^{3 \times 3}$, and $ {\underline{D}_v}_{f_1}(x_4)\in \R^{3 \times 3}$ represent the extended damping, stiffness, and viscous friction matrices, respectively. For the sake of readability, these matrices are provided in Appendix~\ref{app:broken_belt}. Moreover, to observe how the healthy dynamics of the robot is affected by the broken belt fault, the faulty model \eqref{eq:dynamics_state_space_belt}, \eqref{eq:dynamics_state_space_belt2} is introduced, where an additive fault vector $f_1(x_f)$ is incorporated into the healthy dynamics (i.e., in the right-hand side of the last equation in \eqref{eq:dynamics_state_space}), which can be expressed as:
\begin{equation}
    \begin{aligned}
        f_1(x_f) &:= \\ 
        &+\underline{M}(x_2)^{-1} \big( \underline{C}(x_2,x_4) + \underline{D}_v x_4 \\ &+ \underline{D}(x_4 - \mu x_3) + \underline{K}(x_2 - \mu x_1) \big) \\ &- F \underline{M}_{f_1}(x_2,x_{f_1})^{-1} \big( \underline{C}_{f_1}(x_2,x_4,x_{f_1}, x_{f_2} \color{black}) \\ 
        &+ \underline{D}_{f_1}(x_3,x_4) + \underline{K}_{f_1}(x_1,x_2)  + {\underline{D}_v}_{f_1}(x_4) \big)
    \end{aligned}
	\label{eq:broken_belt}
\end{equation}
\end{subequations} 
with $F := \begin{bmatrix} I_2 & {0} \end{bmatrix}$. 

\begin{figure}[tbp]
    \centering
    \includegraphics[width=0.8\linewidth]{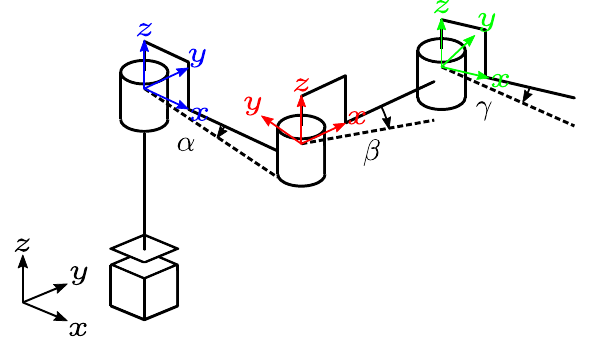}
    \caption{Schematic representation of tilt in the WH robot.}
    \label{fig:tilt_schematic}
\end{figure}

The second fault scenario is tilt in the arms of WH robots. Tilt develops gradually over time, and while this fault does not pose an immediate danger due to its incipient nature, it is still important to detect. Namely, if tilt grows too large (which typically happens on a slow time scale), then the robot will fail. Therefore, if small tilt faults can be detected early, corrective action can be taken before the system fails. Tilt can occur in each arm of the WH robot and is modeled as an additional rotation around the body frame's $y$-axis, as depicted in Figure~\ref{fig:tilt_schematic}. To capture this behavior, the equations of motion are derived with such added rotation, resulting in new inertia and Coriolis matrices that depend on the tilt angles (which are assumed constant in the fault model, given the slow time scale of their growth). This model assumes that the tilt occurs as a rotation around the $y$-axis, followed by a rotation around the $z$-axis for the normal motion of the arms. With some abuse of notation, we write the faulty state as $x_f=\big[ x_1^\top \ \ x_2^\top \ \ x_3^\top \ \ x_4^\top \big]^\top$ for the case of tilt fault (i.e., the meaning of the states is in this case the same as in the healthy case). The equations of motion for the faulty dynamics in the tilt scenario are derived using Euler-Lagrange equations \cite{craig2006introduction} and are used to determine the faulty state-space dynamics, given as

\begin{subequations} \label{eq:faulty_dynamics_tilt}
\begin{equation}
	\begin{aligned}
    \dot{x}_f = &f_{f_2}(x_f,\tau_m;\alpha,\beta,\gamma) \\ 
    = &f_h(x_f,\tau_m) + \begin{bmatrix} 0 & 0 & 0 & I_2\end{bmatrix}^\top f_2(x_f;\alpha,\beta,\gamma),
    \end{aligned}
    \label{eq:dynamics_state_space_tilt}
\end{equation}
where similar to the broken belt model, the faulty dynamics including tilt \eqref{eq:dynamics_state_space_tilt} is introduced, where an additive fault vector $f_2(\cdot)$ affects the healthy dynamics (i.e., the right-hand side of the last equation in \eqref{eq:dynamics_state_space}) as follows:
\begin{equation}
	\begin{aligned}
		&f_2(x_f; \alpha,\beta,\gamma) = \\ &+\underline{M}(x_2)^{-1} \big( \underline{C}(x_2,x_4) + \underline{D}_v x_4 \\ &+ \underline{D}(x_4 - \mu x_3) + \underline{K}(x_2 - \mu x_1) \big) \\  &- \underline{M}_{f_2}(x_2;\alpha,\beta,\gamma)^{-1} \big( \underline{C}_{f_2}(x_2,x_4;\alpha,\beta,\gamma) \\ &+ \underline{D}(x_4 - \mu x_3) + \underline{K}(x_2 - \mu x_1) + \underline{D}_v x_4 \big),
	\end{aligned}
	\label{eq:tilt}
\end{equation}
\color{black}where $\underline{M}_{f_2}(x_2;\alpha,\beta,\gamma)\in \R^{2\times 2}$ and $\underline{C}_{f_2}(x_2,x_4;\alpha,\beta,\gamma) \in \R^{2}$ denote the position-dependent inertia vector and the nonlinear Coriolis and centrifugal matrix of the faulty dynamics for the tilt scenario, respectively, with the included rotations for the tilt angles $\alpha$, $\beta$, and, $\gamma$ for each link (see Figure~\ref{fig:tilt_schematic}). All system matrices for the constant tilt dynamics are given in Appendix~\ref{app:tilt}. 
\end{subequations} 

\subsection{State-Space Formulation of the Healthy Dynamics with Fault Signals}
In a healthy scenario, the system dynamics are governed by Equation \eqref{eq:dynamics_state_space}, \eqref{eq:dynamics_state_space2}. When a fault occurs, the system dynamics switch to those described by Equation \eqref{eq:faulty_dynamics_belt} or Equation \eqref{eq:faulty_dynamics_tilt}, corresponding to the broken belt and tilt fault scenarios, respectively. These dynamics can be utilized to simulate the behavior of both healthy and faulty systems.

For the design of a fault estimation scheme, the known components of the healthy and faulty dynamics could, in principle, be used. However, employing the known components of faulty dynamics (in faulty scenarios) requires the development of a fault detection method to accurately identify when the system transitions into a faulty state. This additional complexity increases the overall challenge of designing a fault diagnosis method.

To simplify the approach, we focus solely on the known components of the healthy dynamics and treat faults as additive unknown signals, irrespective of their internal dynamics\footnote{Note that in the case of the broken belt scenario the fault gives rise to additional second-order dynamics, while this is not the case in the tilt fault scenario.}. With this simplification, the healthy state-space representation in \eqref{eq:dynamics_state_space}, \eqref{eq:dynamics_state_space2} is reformulated to explicitly include fault signals, separating its linear and nonlinear components. This reformulated representation of the faulty dynamics serves as the foundation for the fault estimation scheme proposed in the subsequent sections. To this end, the following model structure is adopted to represent the faulty dynamics:

\begin{subequations} \label{eq:sys}
\begin{equation} 
    \begin{cases}
        \begin{split}
            \dot{x} &= Ax + Bu + S \big(g(x) + f_j\big)  + D_\omega \omega, \quad j\in\{1,2\},\\ 
            y &= Cx + \nu.
        \end{split}
    \end{cases}
    \label{eq:state_space_general}
\end{equation}
where $x=\big[ x_1^\top \ \ x_2^\top \ \ x_3^\top \ \ x_4^\top \big]^\top$, $u = \tau_m$, the measurement noise is denoted by the vector $\nu \in \mathbb{R}^{2}$, the process disturbances are represented by $\omega \in \mathbb{R}^{4}$, capturing modeling uncertainties and external disturbances, and the disturbance distribution matrix is $D_\omega = \begin{bmatrix} 0 & I_4\end{bmatrix}^\top \in \mathbb{R}^{8 \times 4}$. The index $j$ indicates the fault scenario ($j=1$, for the broken belt and $j=2$, for the tilt case). Comparing \eqref{eq:dynamics_state_space}, \eqref{eq:dynamics_state_space2} with \eqref{eq:state_space_general}, the corresponding system matrices in \eqref{eq:state_space_general} are given by 
\begin{equation}
    \begin{aligned}
        A &= \medmath{
        \begin{bmatrix}
            {0} & {0} & I & {0} \\
            {0} & {0} & {0} & I \\
            -\underline{J}^{-1}\underline{K} \mu^2 & \underline{J}^{-1}\underline{K} \mu & -\underline{J}^{-1}\underline{D} \mu^2 & \underline{J}^{-1}\underline{D} \mu \\
            \underline{M}_l^{-1}\underline{K} \mu & -\underline{M}_l^{-1}\underline{K} & \underline{M}_l^{-1}\underline{D} \mu & -\underline{M}_l^{-1}(\underline{D}+\underline{D}_v)
        \end{bmatrix}}, \\
        B &=
        \begin{bmatrix}
            {0} & {0} & \underline{J}^{-1} & {0}
        \end{bmatrix}^\top, \quad
        C = 
        \begin{bmatrix}
            I & {0} & {0} & {0}
        \end{bmatrix}, \\
        S &= 
        \begin{bmatrix}
            {0} & {0} & {0} & I
        \end{bmatrix}^\top, 
        \label{eq:system_matrices}
    \end{aligned}
\end{equation}
where $\underline{M}_l = \underline{M}(x_e)$ denotes the linearized mass matrix at an equilibrium position $x_e$, which corresponds to a relevant operation point of the robot. This linearization is done to capture as much as possible of the nonlinear dynamics in the linear part of the model. The remaining nonlinear part of the dynamics is then given by
\begin{equation}
    \begin{split}
        g(x) &= -\underline{M}(x_2)^{-1} \big( \underline{C}(x_2,x_4) + \underline{D}(x_4 - \mu x_3) \\
        &\quad + \underline{K}(x_2 - \mu x_1) + \underline{D}_v x_4 \big) \\
        &\quad + \underline{M}_l^{-1} \big(\underline{D}(x_4 - \mu x_3) + \underline{K}(x_2 - \mu x_1) \\
        &\quad+ \underline{D}_v x_4  \big).
    \end{split}
    \label{eq:nonlinear_part}
\end{equation}
\end{subequations}
Finally, the fault signal $f_j, j\in\{1,2\}$ is to be estimated as a function of time , $f_j(t)$, as generated by the faulty dynamics in \eqref{eq:faulty_dynamics_belt} or \eqref{eq:faulty_dynamics_tilt}. It is emphasized that, as mentioned at the beginning of this section, the faulty dynamics in \eqref{eq:faulty_dynamics_belt} and \eqref{eq:faulty_dynamics_tilt} should be utilized to generate by simulation the fault signals $f_1(t)$ and $f_2(t)$, along trajectories of the faulty dynamics, for the broken belt and tilt scenarios, respectively.
\color{black}

\section{Model-based Fault Estimation}\label{ch:estimation}
Hereafter, we focus on the faulty state-space dynamics of the WH robot given in \eqref{eq:sys}. For this model, a model-based fault estimation scheme is proposed and later applied to the WH manipulator system in \eqref{eq:sys} to estimate additive fault signals $f_j, j\in\{1,2\}$, for the considered fault scenarios. To implement the fault estimation method, a necessary technical assumption regarding \eqref{eq:sys} is presented below.

\begin{assum}[Regularity]
The following assumptions are required to ensure the regularity of the fault estimation problem as is common in the existing literature \cite{sontag2008input,Keliris2017,han2019intermediate}:
\begin{itemize}
	
	\item 
	\emph{\textbf{State and Input Boundedness:}}  The state $x(t)$ and input $u(t)$ in \eqref{eq:sys} are bounded over any finite time interval.
	\label{assum:state_boundedness} 
	
	\item \emph{\textbf{$\mathcal{C}^r$ Fault and Nonlinearity:}} The function $g(x(t)) + f_j(t), j\in\{1,2\}$, in \eqref{eq:sys} is $r$ times differentiable with respect to $t$.
	\label{assum:cr_assumption}

	\item \emph{\textbf{Disturbance Boundedness:}} The disturbance vectors $\omega(t)$ and $\nu(t)$ in \eqref{eq:sys} are bounded on any finite time interval, and $\nu(t)$ is once differentiable in $t$, i.e., $\dot{\nu}(t)$ exists, and is continuous and bounded over any finite time interval.
	\label{assum:noise_diff}
	
\end{itemize}

\end{assum}

Before introducing the fault estimator in Figure~\ref{ch:fault_estimator}, the following section will present some preliminaries necessary for designing the fault estimator filter.

\subsection{Ultra Local Lumped Fault Representation}
Let us define function $\xi(x(t)) := g(x(t)) + f_j(t))$ in \eqref{eq:sys}, which we will call the lumped fault (sum of the fault an nonlinear terms). Note that $\xi(x(t))$ is an implicit function of time. Hence, an entry-wise $r$-th order Taylor approximation at time $t$ of $\xi$ can be written as $\bar \xi = a_0 + a_1 t+ \dots + a_{r-1} t^{r-1}$ with coefficients $a_i \in {\mathbb{R}^{n_g}}, i= 0, \dots, r-1$. An internal model for this approximation can be written in state-space form as follows:
\begin{equation} \label{eq:fault_model}
	\left\{\begin{aligned}
		\dot{\bar \zeta}_i &= \bar \zeta_{i+1},  \qquad 1 \leq i < r, \\
		\dot{\bar \zeta}_{r} &= {0},\\
		\bar \xi &= \bar \zeta_1,
	\end{aligned}\right.
\end{equation}
where $\bar \zeta_i \in {\mathbb{R}^{n_g}}$. Clearly, in the above model, $\bar \xi^{(r)} = {0}$ holds, which is not the case for the actual signal of the lumped fault $\xi$. Under Assumption~\ref{assum:cr_assumption}, the actual internal state-space representation of $\xi$ is given by
  \begin{equation} \label{eq:fault_system}
  	\left\{\begin{aligned}
  		\dot{\zeta}_i &= \zeta_{i+1}, \qquad 1 \leq i < r, \\
  		\dot{\zeta}_{r} &= \xi^{(r)},\\
  		\xi &= \zeta_1,
  	\end{aligned}\right.
  \end{equation}
where $\zeta_i \in {\mathbb{R}^{n_g}}$. Note that the accuracy of the approximated model \eqref{eq:fault_model} increases as $\xi^{(r)}$ goes to zero (entry-wise). In what follows, to design the fault estimation filter, the system state, $x(t)$,  is augmented with the states of the lumped fault internal state $\zeta_i(t), i \in\left\{1, \ldots, r\right\}$. The system dynamics in \eqref{eq:sys} are extended with \eqref{eq:fault_system}. Then, a linear filter (observer) is designed for the augmented system to simultaneously estimate $x$ and $\zeta_i, i \in\left\{1, \ldots, r\right\}$ using model \eqref{eq:fault_model}. 

\subsection{Augmented Dynamics}
Based on the lumped fault $\xi$ introduced above, define the augmented state $x_a:=(x, \zeta_1,\zeta_2,\ldots,  \zeta_{r})$, and rewrite the augmented dynamics using \eqref{eq:sys} and \eqref{eq:fault_system} as
\begin{subequations} 		\label{eq:augmented}
    \begin{equation}\label{eq:augmented_system}
        \begin{aligned}
            \left\{\begin{aligned}
                \dot{x}_{a} &=A_{a} x_{a}+B_{a} u +D_{a} \omega_{a},\\
                y &= C_{a} x_{a}+\nu,
            \end{aligned}\right.
        \end{aligned}
    \end{equation}
    \begin{equation} 		\label{eq:augmented_matrices}
    \begin{aligned}
    A_a &:= 
    \begin{bmatrix}
        A & S & {0} \\
        {0} & {0} & I_{d_n} \\
        {0} & {0} & {0} 
    \end{bmatrix}, \quad
    B_a := 
    \begin{bmatrix}
        B \\ {0}
    \end{bmatrix}, \\
    D_a &:= 
    \begin{bmatrix}
        D_\omega & {0} \\
        {0} & {0} \\
        {0} & I_{n_g}
    \end{bmatrix}, \quad 
    \omega_{a} :=	
    \begin{bmatrix}
    		\omega \\
        \xi^{(r)} 
    \end{bmatrix} \\
    C_a &:= 
    \begin{bmatrix}
        C & {0}
    \end{bmatrix},
\end{aligned}
    \end{equation}
\end{subequations}
with $d_n = (r-1)n_{g}$.

\subsection{Fault Estimator}\label{ch:fault_estimator}
In this section, inspired by observer-based methods, the fault estimator filter is proposed as
\begin{subequations} \label{eq:observer}
    \begin{align}
        \dot{z} &= N z+G u+L y, \label{eq:observer_dynamics} \\ 
        \hat{f} &= \bar{C}(z-Ey)-g\big(V_a (z-Ey) \big), \label{eq:observer_dynamics_phi}
    \end{align}
    with filter state $z \in {\mathbb{R}^{n_z}}$, $n_z = n+r n_{g}$, 
\begin{equation*} 	
        \bar{C} := {\left[\begin{array}{ccc}
                {0} & I_{n_{g}} & {0} 
            \end{array}\right]}, \quad
        V_a := {\left[\begin{array}{cc}
                I_n & {0}
            \end{array}\right]},
    \end{equation*}
and matrices $\{N,G,L\}$ defined as
    \begin{equation}		\label{eq:observer_matrices}
        \begin{aligned}
            N &:=M A_a-K C_a,\quad M:=I+E C_a, \\
            G&:=M B_a,\quad L:=K(I+C_a E)-M A_a E.
        \end{aligned}
    \end{equation}
\end{subequations}
Matrices $E$ and $K$ are filter gains to be designed. Note that it is assumed that the fault-free system dynamics are known. Hence, the nonlinear function $g(\cdot)$, can be used to extract fault signals algebraically as in \eqref{eq:observer_dynamics_phi}. According to \eqref{eq:observer_dynamics_phi}, the part of the augmented state, $x_a$, that is used to reconstruct fault signal is $\bar{C}_a x_{a}$ with
\begin{equation} 	\label{eq:c_bar_a}
    \bar{C}_a := {\left[\begin{array}{cc}
            V_a^\top &
            \bar{C}^\top 
        \end{array}\right]^\top}.
\end{equation}
In the following section, to provide a problem for synthesis of the fault estimator filter (i.e., selection of filter gains $\psi := \{E,K\}$), the fault estimator error dynamics are analyzed.
\subsection{Fault Estimator Error Dynamics} \label{sec:error_dynamics}
Consider the augmented state estimate $\hat{x}_a$ and define estimation error as 
\begin{equation*}
    e:=\hat{x}_a-x_a=z-x_a-E y=z-M x_a-E\nu.
\end{equation*}
Then, given the algebraic relations in \eqref{eq:observer_matrices}, the estimation error dynamics can be written as
\begin{equation*}
    \begin{aligned}
        \dot{e} &= Ne - M D_a \omega_{a} + {\left[\begin{array}{cc}
                K & -E
            \end{array}\right]} {\left[\begin{array}{cc}
                \nu \\ \dot{\nu}
            \end{array}\right]}.
    \end{aligned}
\end{equation*}
Define $\nu_a := (\nu, \dot{\nu})$, $e_d := \bar{C}_a e$ with $\bar{C}_a$ as in \eqref{eq:c_bar_a}, and
$
    \bar{B} := {\left[\begin{array}{cc}
            K & -E
        \end{array}\right]}.
    \label{eq:b_bar}
$
Then, the estimation error dynamics is given by
\begin{equation}
    \left\{\begin{aligned}
        \dot{e} &=Ne - M D_a \omega_{a} + \bar{B} \nu_a, \\
        e_d &= \bar{C}_a e.
    \end{aligned}\right.
    \label{eq:error_system}
\end{equation}
Define the transfer function matrices
\begin{equation}
    \begin{aligned}
        T_{e_d \omega_a}(s) &:= -\bar{C}_a(s I-N)^{-1} M D_a, \\
        T_{e_d\nu_a}(s) &:= \bar{C}_a(s I-N)^{-1} \bar{B},
    \end{aligned}
    \label{eq:tfs}
\end{equation}
where $s \in \mathbb{C}$, $T_{e_d \omega_a}(s)$ and $T_{e_d\nu_a}(s)$ denote the corresponding transfer function matrices from inputs $\omega_a$ and $\nu_a$, both to the output $e_d$, respectively. In the context of fault estimation, the estimation error $e_d$ is especially important, as it represents not only the error in the faults to-be-estimated introduced into the system but also the error in the state estimates, which serve as arguments for the nonlinearity in \eqref{eq:observer_dynamics_phi}.
Now, the fault estimator synthesis problem for the WH system in \eqref{eq:sys} can be stated as follows.
 
\begin{prob}\emph{\textbf{(Fault Estimator Synthesis)}}
\label{prob:fault_estimation}
Consider the faulty WH system dynamics in \eqref{eq:sys} with known input and output signals, $u(t)$ and $y(t)$. Further, consider the internal lumped fault dynamics \eqref{eq:fault_system}, its Taylor approximation \eqref{eq:fault_model}, the augmented dynamics \eqref{eq:augmented}, the fault estimator in \eqref{eq:observer}, and the transfer matrices in \eqref{eq:tfs}. Design the filter gain matrices $\psi = \{E,K\}$ such that, under Assumption~\ref{assum:state_boundedness}, we have: \\
\textbf{\emph{1) Stability:}} The estimation error dynamics \eqref{eq:error_system} is input-to-state stable (ISS) with respect to input $\bar \omega_a = (\omega_a, \nu_a)$\emph{;} \\
    \emph{\textbf{2) Disturbance Attenuation:}} 
\begin{equation} \label{eq:J1}
        J_1(\psi) := \|T_{e_d \omega_a}\|_{\infty} = \sup _{\mu \in \mathbb{R}^+} \sigma_{\max }(T_{e_d \omega_a}(i \mu)) \leq \bar \lambda,
\end{equation}
 is bounded by some known $\bar \lambda > 0$\emph{;} \\[1 mm]
\emph{\textbf{3) Noise Rejection:}} 
\begin{equation} \label{eq:J2}
    \begin{aligned}
        J_2(\psi) &:= \|T_{e_d \nu_a}\|_{H_2} \\
           &= \sqrt{\frac{1}{2 \pi} \operatorname{trace} \int_{-\infty}^{\infty} T_{e_d\nu_a}(i \mu) T_{e_d\nu_a}^{H}(i \mu) \mathrm{~d} \mu} \leq \bar \gamma,
    \end{aligned}
\end{equation}
is bounded by some known $\bar \gamma > 0$\emph{.}
\end{prob}
Under Assumption~\ref{assum:state_boundedness}, the essence of Problem~\ref{prob:fault_estimation} is to find a fault estimator that: ensures a bounded estimation error, $e(t)$; the $H_{\infty}$-norm of $T_{e_d \omega_a}(s)$, the transfer matrix from external disturbances and lumped fault model mismatch to the performance output $\bar{C}_a e$ is upper bounded by $\bar \lambda$; the $H_2$-norm of $T_{e_d \nu_a}(s)$, the transfer matrix from measurement noise to the performance output, is upper bounded by $\bar \gamma$.
 
\subsection{Fault Estimator Synthesis}
In the following theorem, the solution to Problem~\ref{prob:fault_estimation} is given in terms of the solution of semi-definite program, where the aim is to minimize the $H_{\infty}$-norm of $T_{e_d \omega_a}(s)$ for an acceptable upper bound on the $H_2$-norm of $T_{e_d \nu_a}(s)$. Moreover, the input-to-state stability (ISS) constraint is added to this program to enforce the internal stability of the estimation error dynamics. 

\begin{thm}\emph{\textbf{(Fault Estimator Synthesis)}
Consider the augmented dynamics \eqref{eq:augmented}, the fault estimator filter as defined in \eqref{eq:observer}, the corresponding estimation error dynamics \eqref{eq:error_system}, and the transfer functions \eqref{eq:tfs}. To design parameters of the optimal mixed $H_2/H_{\infty}$ fault estimators, solve the following convex program:
\begin{subequations} \label{eq:mimization}
        \begin{equation}
            \begin{array}{cl}
                \min \limits_{P, R, Q, Z, \bar \lambda, \bar \gamma}  & \bar \lambda \\
                \text{s.t.} & 						\vspace{1 mm}
                X 	 +  \epsilon I \preceq 0, \\
                &\left[\begin{array}{ccc}
                    X & Y & \bar{C}_a^{\top} \\
                    * & -\bar \lambda I  & {0} \\
                    * & * & -\bar \lambda I
                \end{array}\right] \prec 0,\\[5mm]
                & \left[\begin{array}{ccc}
                    X &  Q & -R\\
                    * & -\bar \gamma I & {0} \\
                    * & * & -\bar \gamma I
                \end{array}\right] \prec 0, \\[3mm]
                &\left[\begin{array}{cc}
                    P & \bar{C}_a^{\top} \\
                    * & Z
                \end{array}\right] \succ 0, \quad P \succ 0, \\[3mm] 
                &\operatorname{trace}(Z)<\bar \gamma, \quad \bar \gamma \leq \bar \gamma_{max}, \\  &\bar \lambda, \bar \gamma  > 0
            \end{array}
        \end{equation} 
        with 
			\begin{equation}
				\begin{aligned}
					X :=&A_{a}^{\top}  P +A_{a}^{\top} C_{a}^{\top} R^{\top}-C_{a}^\top  Q^\top + P A_{a} +R C_{a} A_{a} \\
                        &-  Q C_{a}, \\ Y :=& -(P + R C_a) D_a,
				\end{aligned}
				\label{eq:X}
			\end{equation}
   \end{subequations}
given $\epsilon, \bar \gamma_{max} > 0$, $\bar{C}_a$ in \eqref{eq:c_bar_a}, and the remaining matrices as defined in \eqref{eq:augmented_matrices}. Denote the resulting optimizers as $P^\star, R^\star, Q^\star, Z^\star, \bar \lambda^\star$, and $\bar \gamma^\star$. Then, the parameters of the estimation filter in \eqref{eq:observer}, given by $\psi = \psi^\star = \{ E^\star = P^{\star^{-1}} R^\star, K^\star = P^{\star^{-1}} Q^\star \}$, guarantee the following properties:
        \begin{enumerate}
            \item The estimation error dynamics in \eqref{eq:error_system} is ISS with ISS-gain from $(\omega_a, \nu_a)$ to the estimation error upper bounded by $ 2 \|P^\star {\left[\begin{array}{ccc} (I+E^\star C_a) D_a & -K^\star & E^\star \end{array}\right]} \| \epsilon^{-1}$.
            \item $J_1(\cdot)$ in \eqref{eq:J1} is upper bounded by $\bar \lambda^\star$, i.e., \linebreak 
$\|T_{e_d \omega_a}\|_{\infty}<\bar \lambda^\star$.
            \item $J_2(\cdot)$ in \eqref{eq:J2} is upper bounded by $ \bar \gamma^\star$, i.e., \linebreak  $	\|T_{e_d \nu_a}\|_{H_2}<\bar \gamma^\star$.
    \end{enumerate}}
    \label{theorem:optimal_estimator}
\end{thm}
\emph{\textbf{Proof}:} 
The proof can be found in Appendix~\ref{ap:thm1_proof}.
\hfill $\blacksquare$

\subsection{Simulation-Based Case Study Results on Fault Estimation}
\begin{figure}[tbp]
    \centering
    \includegraphics[width=1\linewidth]{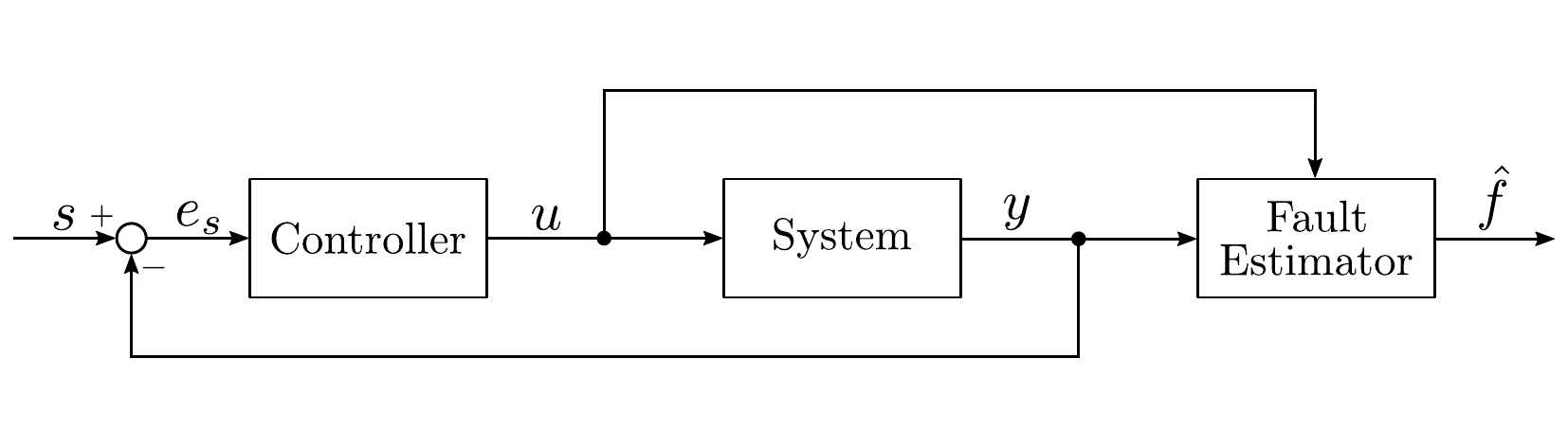}
    \caption{Schematic overview of the model-based fault estimation scheme.}
    \label{fig:fault_estimation}
\end{figure}

The performance of the fault estimator is evaluated in a simulation environment where faults are injected into the wafer handler robot model at a certain time instance. The fault estimator is applied to both the broken belt and tilt fault scenarios. A schematic overview of the considered simulation environment for model-based fault estimation is given in Figure~\ref{fig:fault_estimation}.
In this figure, $s$ is the provided trajectory setpoint for the motor angles $\theta_{m_1}$ and $\theta_{m_2}$, $e_s$ denotes the setpoint error, $u$ is the supplied control input provided by the closed loop controller as given in Appendix \ref{ap:controller}\color{black}, $y$ is the system output, and $\hat{f}$ denotes the estimated fault.
The considered measurement noise and fault parameters for the simulations are given in Table~\ref{tab:estimation_parameters_noise}. Here, the simulation setting is based on the expected behavior of the system in practice. 

Regarding the fault parameters, for tilt, the fault parameters $\alpha$, $\beta$, and $\gamma$ describe the specific tilt angles occurring in each link of the manipulator, respectively (see Section~\ref{ch:modeling} for further details). For a broken belt, we switch from the healthy model to a faulty model with a higher dimension. The additional states of the faulty model, corresponding to the position and velocity of the end effector (i.e., ${\theta}_{a_3}$ and $\dot{\theta}_{a_3}$, respectively), are initialized using their healthy constrained values, (i.e., $({\theta}_{a_1}+{\theta}_{a_2})/2$ and $(\dot{\theta}_{a_1}+\dot{\theta}_{a_2})/2$, respectively). These values vary depending on the time of fault occurrence. Thus, the only fault parameter for a broken belt is the fault occurrence time.

\begin{table}[tbp]
\centering
\caption{Fault parameters used for fault estimation.}
\label{tab:estimation_parameters_noise}
\begingroup
\setlength{\tabcolsep}{5pt}
\setlength\extrarowheight{2pt}
\begin{tabular}{llll}
\hline
Parameters                     & Symbol         & Value                       & Unit \\ \hline
Trajectory setpoint $\theta_a$ & $s$              & $\begin{bmatrix}
4 + 2 \sin\left(\frac{1}{2}t\right) \\
1.5 + 2 \sin\left(\frac{1}{2}t\right)
\end{bmatrix}$  & rad \\[3 mm]
Noise amplitude                & $\nu$          & $\mathcal{U}(-1,1)\cdot10^{-6}$          & -    \\ \hline
\multicolumn{4}{l}{Fault Parameters Broken Belt}                          \\ \hline
Time of fault occurrence       & - & 50                          & $s$  \\ \hline
\multicolumn{4}{l}{Fault Parameters Tilt}                                 \\ \hline
Upper arm tilt                 & $\alpha$ & $5$             & deg  \\
Lower arm tilt                 & $\beta$  & $5$             & deg  \\
End-effector tilt              & $\gamma$ & $5$             & deg  \\ \hline
\end{tabular}
\endgroup
\end{table}

\begin{figure}[tbp]
    \centering
    \includegraphics[width=\linewidth]{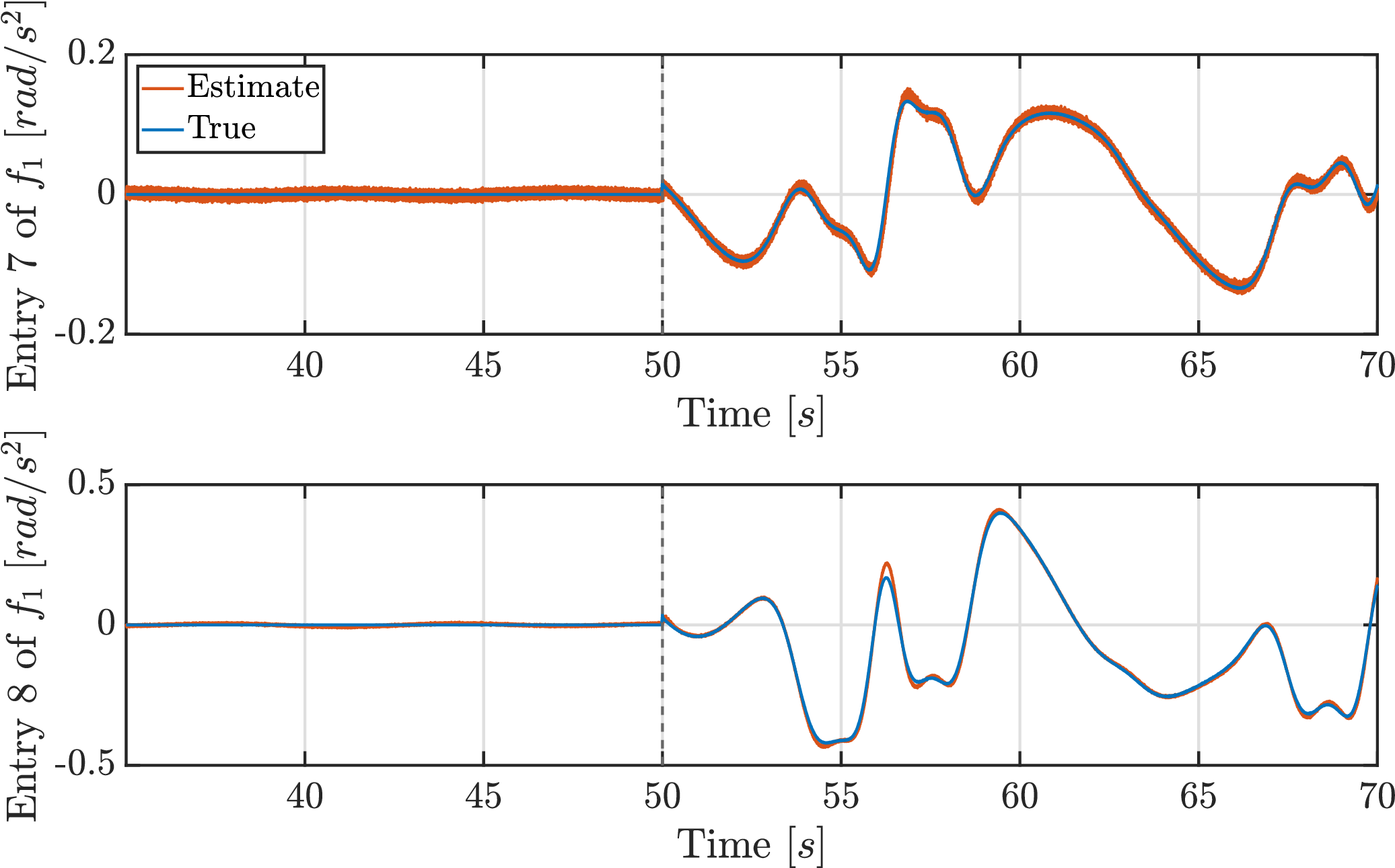}
    \caption{Fault estimation for a broken belt at $t=50$ s with measurement noise.}
    \label{fig:belt_noise}
\end{figure}

\color{black}
The results for fault estimation with measurement noise for a broken belt are depicted in Figure~\ref{fig:belt_noise}. It is clear that after the injection of the fault at $t=50$ s, the estimated signal accurately follows the effect the true fault.   




\begin{figure}[t]
    \centering
    \includegraphics[width=\linewidth]{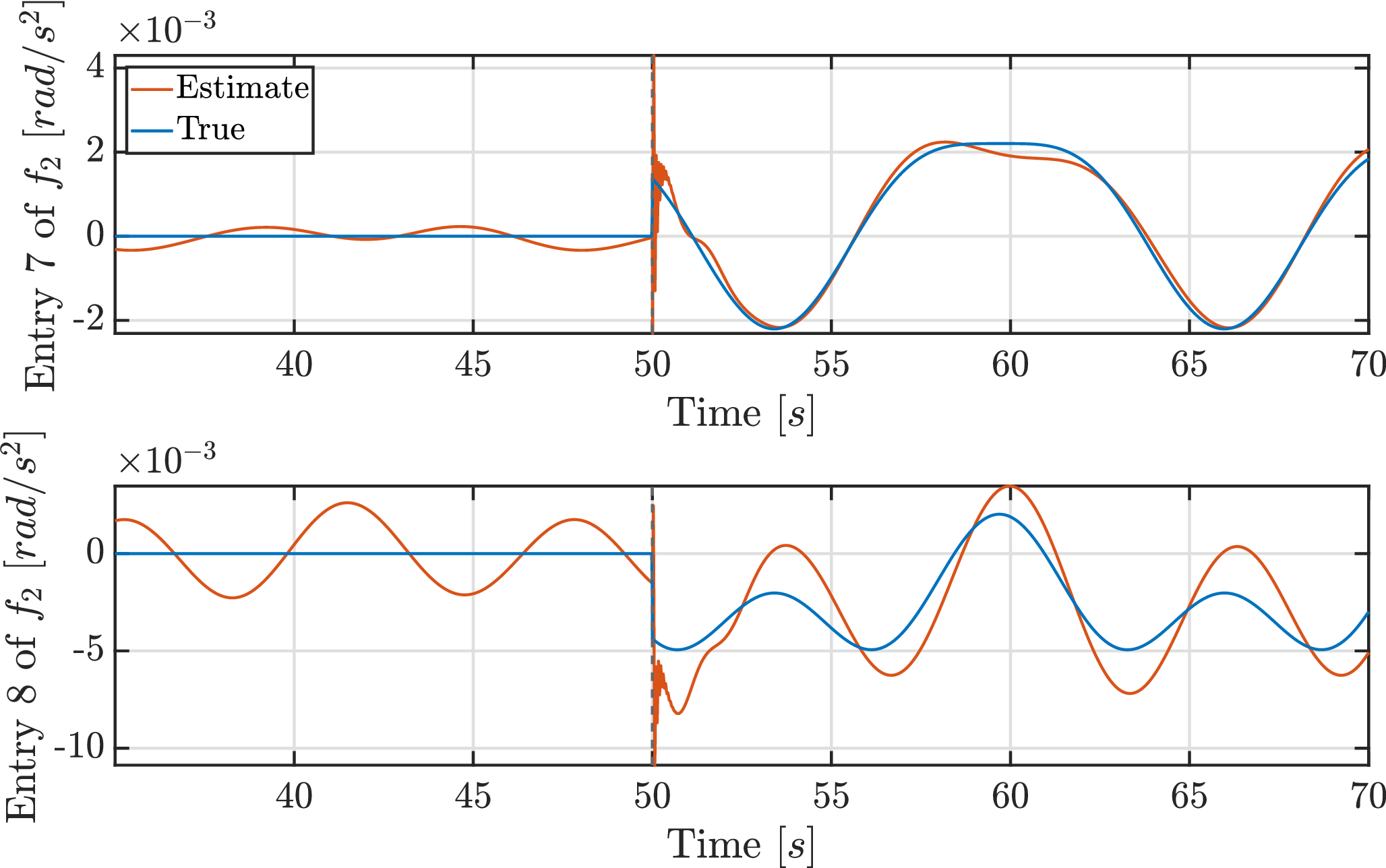}
    \caption{Fault estimation for tilt at $t=50$ s without measurement noise.}
    \label{fig:tilt}
\end{figure}

The results for fault estimation without measurement noise in the case of tilt are given in Figure~\ref{fig:tilt}. Note that the effect of tilt on the healthy dynamics compared to the broken belt fault is significantly smaller (see the fault magnitudes in Figures~\ref{fig:belt_noise} and \ref{fig:tilt}), making its estimation challenging. However, after the fault is injected at $t=50$ s, the proposed fault estimator is able to approximately estimate the fault signal. Furthermore, it should be noted that in the healthy operation mode, the fault estimate is not exactly equal to zero, as it would be in the ideal case. This behavior results from the nonlinearities in the system that cannot be modeled exactly using the ultra-local model in \eqref{eq:fault_system}. While the fault estimate does not converge to zero in the healthy operation mode, it does converge to a bounded estimation error, which is minimized in the $H_2/H_{\infty}$ sense as considered in \eqref{eq:mimization}. Moreover, even for tilt angles as low as $2$ deg, the fault estimator can generate a signature indicating fault occurrence, which could be used for fault isolation. For extremely small tilt angles (on the order of $1$ mrad) since the fault estimate is not zero for the healthy case in the proposed method, the fault estimate is not distinguishable before and after fault occurrence. To address this challenge, and based on the insights obtained here, we propose to exploit additional sensors able to directly measure effects of the dynamics in z-direction, since the tilt more directly affects those dynamics than the planar dynamics, in which the tilt is only a higher-order effect.


\begin{figure}[t]
    \centering
    \includegraphics[width=\linewidth]{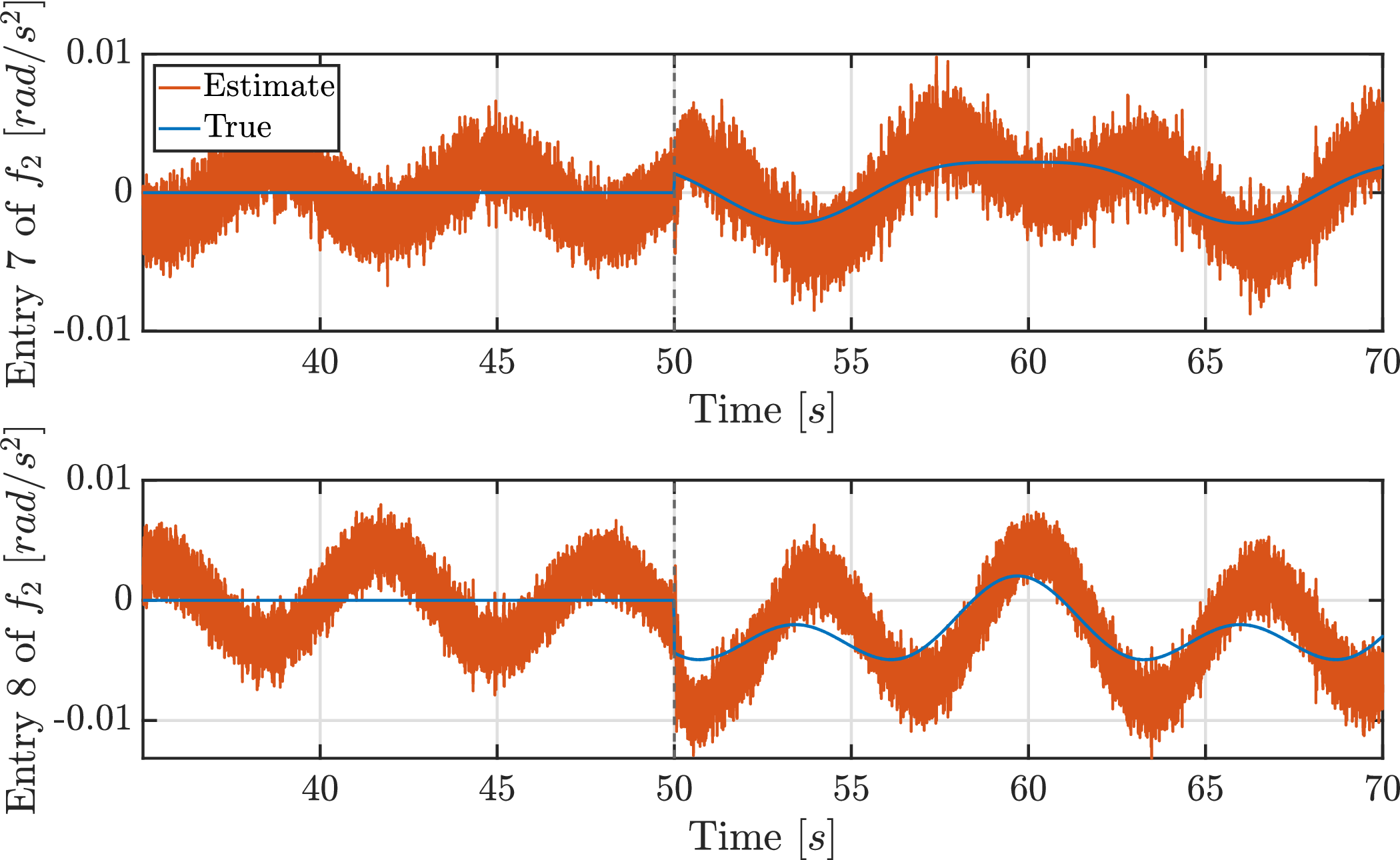}
    \caption{Fault estimation for tilt at $t=50$ s  with measurement noise.}
    \label{fig:tilt_noise}
\end{figure}

When including measurement noise into the simulation, similar estimation performance can be noticed, see Figure~\ref{fig:tilt_noise}. It becomes clear that in both the healthy and faulty operation modes, the measurement noise affects the bound of the estimation error, resulting in poor performance. This poor performance can be attributed to the fact that tilt is a very small fault, which makes it challenging to estimate, especially in the case where measurement noise is present. However, the estimated fault signal still shows different behavior between the healthy and the faulty operation modes. Due to these differences, the fault estimates can still be supplied to a data-driven classifier for fault isolation.

\section{Data-based Fault Detection and Isolation}\label{ch:hybrid}
In the previous section, a model-based fault estimation filter has been introduced that estimates time-varying fault estimates for two representative fault scenarios for the wafer handler robot. Now, it is still an open question how to decide, based on these estimated fault signals: 1) when a fault occurs or not (detection) and 2) which fault occurs (isolation). Several classification algorithms can be considered for data-driven health mode classification (fault detection and isolation). Synthetic data, either healthy or faulty, has been generated using a model of the system. This synthetic data can be fed into these classifiers which, in turn, evaluate the data and attribute it to a certain class. In this section, the supervised classification method of support vector machines (SVM) \cite{deisenroth2020mathematics} is considered for fault isolation.

\subsection{Data Generation}
To train the classification models for fault isolation, first labeled training data needs to be collected for the different fault scenarios. To this end, the model-based fault estimator given in Section~\ref{ch:estimation} is employed to generate synthetic data of the faults. In practice, the trained classifier can be fine-tuned by the data from the actual system using transfer learning ideas \cite{pan2009survey}. Faulty data is collected from the time instance that the fault is injected into the model until 30 seconds after the fault injection. Moreover, since the faults may occur in slightly different configurations, the fault data is collected for different combinations of the fault parameters. Finally, to increase the robustness of the classifiers, measurement noise is added while generating the fault data. Alongside the fault data, healthy data will also be generated such the classifiers will be able to differentiate between a healthy and faulty scenarios.


Eventually, to evaluate the performance of the classifiers, test data is required. This test data is again different from the training data, but it also differs from the validation data. In this way, the classifier is supplied with data it has not seen before, to be able to objectively evaluate the performance of the classifier.

The fault parameters considered for generating the training and test data are given in Tables~\ref{tab:training_parameters} and \ref{tab:test_parameters}, respectively. For both faults, broken belt and tilt, each possible combination of the listed parameters in Tables~\ref{tab:training_parameters} and \ref{tab:test_parameters} is used to obtain extensive data sets that can be used for training and validating the classification models for fault isolation. It is important to note that this is done for each fault individually, this means that the situation where a broken belt and tilt occur simultaneously is not considered in this paper.

\begin{table}[tbp]
\centering
\caption{Fault parameters used for training.}
\label{tab:training_parameters}
\begingroup
\setlength{\tabcolsep}{5pt}
\setlength\extrarowheight{2pt}
\begin{tabular}{llll}
\hline
Parameters                     & Symbol         & Value                       & Unit \\ \hline
Trajectory setpoint $\theta_a$ & $s$              & $\begin{bmatrix}
4 + 2 \sin\left(\frac{1}{2}t\right) \\
1.5 + 2 \sin\left(\frac{1}{2}t\right)
\end{bmatrix}$  & rad \\[3 mm]
Noise amplitude                & $\nu$          & $\mathcal{U}(-1,1)\cdot10^{-6}$          & -    \\ \hline
\multicolumn{4}{l}{Fault Parameters Broken Belt}                          \\ \hline
Time of fault occurrence       & - & $30, 35, 40, \ldots, 65$                          & $s$  \\ \hline
\multicolumn{4}{l}{Fault Parameters Tilt}                                 \\ \hline
Upper arm tilt                 & $\alpha$ & $2, 5$             & deg  \\
Lower arm tilt                 & $\beta$  & $2, 5$             & deg  \\
End-effector tilt              & $\gamma$ & $2, 5$             & deg  \\ \hline
\end{tabular}
\endgroup
\end{table}


\begin{table}[tbp]
\centering
\caption{Fault parameters used for testing.}
\label{tab:test_parameters}
\begingroup
\setlength{\tabcolsep}{5pt}
\setlength\extrarowheight{2pt}
\begin{tabular}{llll}
\hline
Parameters                     & Symbol         & Value                       & Unit \\ \hline
Trajectory setpoint $\theta_a$ & $s$              & $\begin{bmatrix}
4 + 2 \sin\left(\frac{1}{2}t\right) \\
1.5 + 2 \sin\left(\frac{1}{2}t\right)
\end{bmatrix}$  & rad \\[3 mm]
Noise amplitude                & $\nu$          & $\mathcal{U}(-1,1)\cdot10^{-6}$          & -    \\ \hline
\multicolumn{4}{l}{Fault Parameters Broken Belt}                          \\ \hline
Time of fault occurrence       & - & $27, 31.5, 36, \ldots, 58.5$                          & $s$  \\ \hline
\multicolumn{4}{l}{Fault Parameters Tilt}                                 \\ \hline
Upper arm tilt                 & $\alpha$ & $1.8, 4.5$             & deg  \\
Lower arm tilt                 & $\beta$  & $1.8, 4.5$             & deg  \\
End-effector tilt              & $\gamma$ & $1.8, 4.5$             & deg  \\ \hline
\end{tabular}
\endgroup
\end{table}

\subsection{Validation of Classification Models}
To evaluate the performance of the classification models and to compare the different methods with each other, performance measures will be introduced that describe the accuracy of a trained model. A straightforward starting point for evaluating the general accuracy is by calculating the correctly predicted data points as a percentage of the total evaluated data points. The predicted data points can be attributed to one of the following categories; true positive (TP), true negative (TN), false positive (FP), and false negative (FN). Correctly predicted data points will fall in either the true positive or true negative category, while incorrectly predicted data points will fall in the false positive or false negative category. Here, the positive classification refers to the fact that a fault is present in the system, whereas the negative refers to the healthy situation of the system. While it is interesting to see how often a prediction is done correctly, it is at least as (if not more) important to evaluate how often a prediction is done incorrectly. If, for example, a fault is present but the system is classified as healthy, the system could be subjected to severe damage when the system continues to operate with an ongoing fault. On the other hand, a system suffers from unnecessary downtime if the system is classified as faulty when it is actually healthy. Hence, performance measures must be introduced to relate the overall accuracy of the FDIE scheme. These measures are defined as the fault detection rate (FDR) and the false alarm rate (FAR) given by
\begin{equation}
    FDR := \frac{TP}{TP + FN}, \quad FAR := \frac{FP}{TN + FP}.
    \label{eq:FDR}
\end{equation}
An overall accuracy measure can be obtained by simply computing the ratio of the number of data points that are predicted correctly, this total detection rate (TDR) is defined as
\begin{equation}
    TDR := \frac{TP + TN}{TP + TN + FP + FN}.
    \label{eq:TDR}
\end{equation}
This accuracy gives a global impression of the detectability of the FDIE scheme regarding fault detection. However, it may not always reflect the performance for specific situations regarding fault isolation. For example, it could be the case that for most classes the classifier is able to identify the class correctly, but it completely misclassifies one. In such a scenario, the TDR measure may still reflect a high accuracy for the model while in reality, its performance may not be useful. To include the effect of poor detectability of certain types of fault, i.e., fault isolability, the harmonic mean accuracy (HMA) measure is proposed and is given as
\begin{equation}
    HMA := \frac{N_c}{\sum_{n_c=1}^{N_c} TDR_{n_c}^{-1}},
    \label{eq:HMA}
\end{equation}
where $N_c$ is the total number of classes (i.e., $N_c = 3$ for the considered system, two fault scenarios and healthy mode) and $TDR_{n_c}$ corresponds to the total detection rate for a class $n_c$. This measure considers the TDR as introduced in \eqref{eq:TDR} for each individual class $n_c$.

Subsequently, the performance measures given in \eqref{eq:TDR} and \eqref{eq:HMA} are used to assess the accuracy of the trained models based on their predictions on the validation data.

\subsection{Results for Test Data}
\begin{figure}[tbp]
    \centering
    \includegraphics[width=\linewidth]{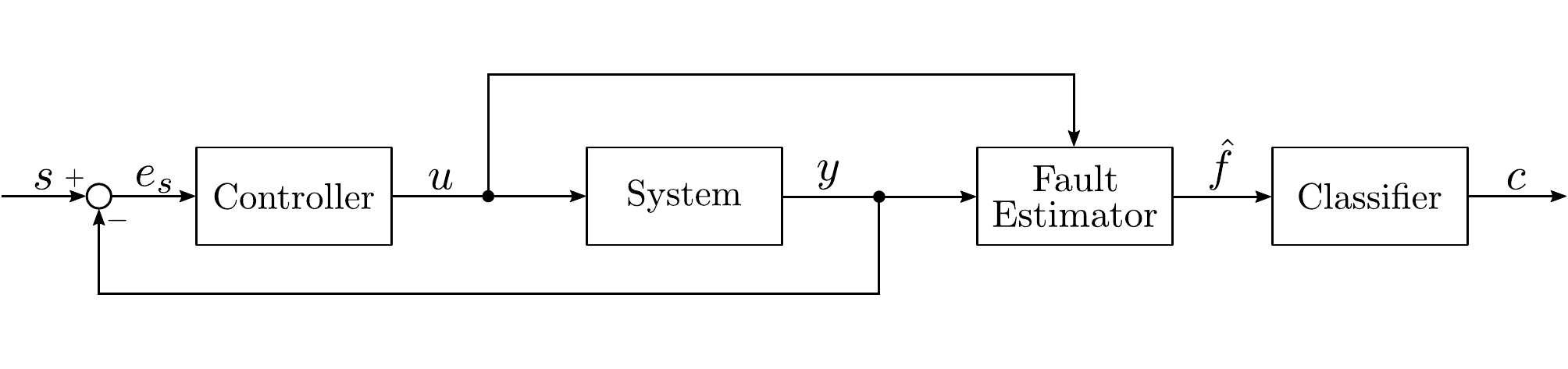}
    \caption{Schematic overview of the hybrid FDIE scheme.}
    \label{fig:hybrid_FDIE}
\end{figure}

\begin{figure}[tbp]
    \centering
    \includegraphics[width=\linewidth]{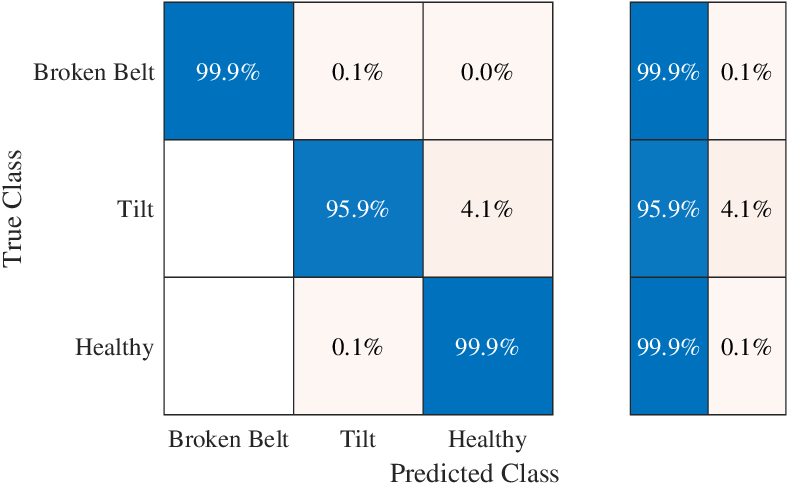}
    \caption{SVM confusion chart for the fault estimate (i.e., the proposed method) with optimized hyperparameters.}
    \label{fig:fault_confusion_svm_hyp_opt}
\end{figure}

To evaluate the method's performance on the test data, the fault parameters are provided in Table~\ref{tab:test_parameters}. Additionally, the accuracy measures, TDR and HMA, as defined in \eqref{eq:TDR} and \eqref{eq:HMA}, respectively, are used to assess the performance on the test data. For analyzing the FDIE scheme's performance, the fault estimation scheme is updated according to the schematic overview shown in Figure~\ref{fig:hybrid_FDIE}. For the proposed FDIE scheme, a TDR performance of $98.57 \%$ is achieved. Furthermore, a HMA performance of $98.53\%$ is achieved, underscoring the importance of HMA in providing better insight into overall fault isolation performance. To further evaluate the classifiers' performance, confusion charts are used, showing the isolation performance of the classifiers for each fault scenario. The confusion chart for the fault estimates, including measurement noise, is shown in Figure~\ref{fig:fault_confusion_svm_hyp_opt}. Overall, it can be concluded that the proposed hybrid approach for fault diagnosis demonstrates strong performance in detecting and isolating faults in the WH robots for the considered fault scenarios.

\begin{figure}[tbp]
    \centering
    \includegraphics[width=\linewidth]{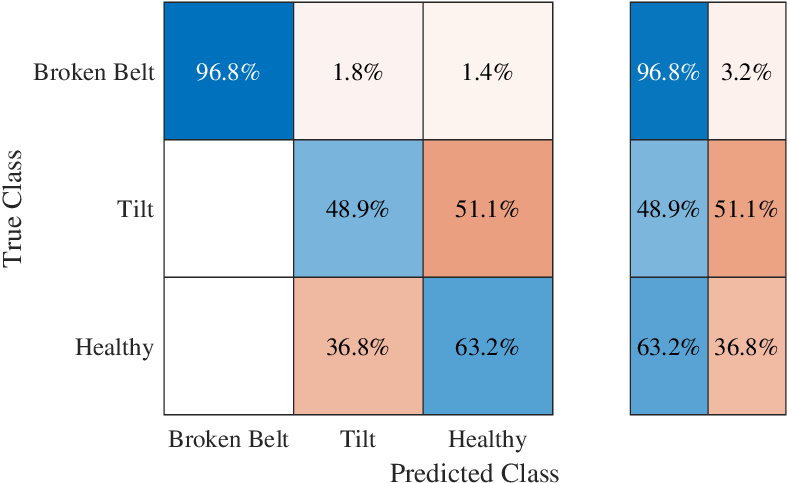}
    \caption{SVM confusion chart for the raw system input-output data with optimized hyperparameters.}
    \label{fig:output_confusion_svm}
\end{figure}

For the sake of comparison, purely data-driven fault isolation is also considered, where the system input $u$ and output $y$ data (instead of the fault estimate $\hat{f}$ in the proposed method) is directly fed into the SVM-based classifiers for fault isolation, establishing a benchmark. The fault isolation models are trained using the raw input-output data of the system. In this case, the overall TDR performance reached only $69.65\% $, clearly indicating that the classifier's performance is significantly lower when relying solely on input-output data compared to using fault estimates for classification. A similar conclusion can be drawn from the HMA performance, which achieved just $64.39\%$. It is evident that the classifier's performance is significantly worse in this case. Examining the confusion chart in Figure~\ref{fig:output_confusion_svm} further highlights the poor performance on raw output data. It is worth pointing out that, since the broken belt fault is a large fault, it can be estimated with acceptable performance even using the raw input-output data. This highlights that the proposed method can enhance fault isolation performance for relatively smaller faults in WH robots.

Ultimately, comparing the results of fault isolation using raw system data to those that use fault estimates generated from a model-based fault estimator underscores the advantage of the proposed hybrid approach. The fault estimate signals, provided by the model-based estimator, offer much more information about the fault type than the raw output signal, demonstrating the clear benefit of the hybrid method over a purely data-driven FDI approach for the WH robots.

\section{Conclusion}\label{ch:conclusion}
In this paper, a hybrid model- and data-based fault detection, isolation, and estimation (FDIE) scheme for wafer handler (WH) robots has been developed. The proposed model-based component of the FDIE algorithm consists of an observer-based fault estimator. This fault estimator is employed to provide an estimate of the fault, given the input and the measured output of the system. Subsequently, this fault estimate is used by the data-driven part of the FDIE scheme. The considered data-driven component of the FDIE scheme consists of a machine learning algorithm for classification. Hereto, a support vector machine (SVM) is proposed. This classifier is then employed to identify and isolate the different types of fault, based on the supplied fault estimate. The hybrid FDIE algorithm is then applied to two case studies that represent two fault scenarios occurring to WH robots deployed in the semiconductor manufacturing field. The results of simulation-based case studies show that the hybrid FDIE algorithm outperforms the solely data-based approach. Future work could include improving fault estimation for extremely small tilt angles by incorporating additional sensor measurements. This adjustment could potentially enhance the accuracy of fault estimation for extremely small tilt angles.

\section*{Acknowledgments}
This publication is part of the project Digital Twin project 4.3 with project number P18-03 of the research programme Perspectief which is (mainly) financed by the Dutch Research Council (NWO).

\appendix
\section{Appendices}

\subsection{Matrix Entries for Equations of Motion}\label{app:eom}
The complete matrices describing the equations of motion (EOM) as used in \eqref{eq:dynamics_complete} are given in this section. The constant matrices included in the EOM are given by
\begin{align*}
    \begin{aligned}
        \underline{J} &:= 
        \begin{bmatrix}
            J_m & 0 \\
            0 & J_m
        \end{bmatrix}, \\
        \underline{D} &:= 
        \begin{bmatrix}
            d_{r_1} & 0 \\
            0 & d_{r_2}
        \end{bmatrix},
    \end{aligned}
    \quad
    \begin{aligned}
        \underline{K} &:= 
        \begin{bmatrix}
            c_{r_1} & 0 \\
            0 & c_{r_2}
        \end{bmatrix}, \\
        \underline{D}_v &:= 
        \begin{bmatrix}
            d_v & 0 \\
            0 & d_v
        \end{bmatrix},
    \end{aligned}
\end{align*}
where $J_m$ is the inertia of the motors, $d_{r_i}$ with $i\in \{1,2\}$ denotes the relative damping of each power transmission, $c_{r_i}$ with $i\in \{1,2\}$ describes the relative stiffness of each power transmission and $d_v$ is the viscous friction acting on the arms of the robot. Subsequently, the nonlinear terms of the EOM are given below. First, the nonlinear inertia matrix is defined as
\begin{equation*}
    \underline{M}(\theta_a) := 
    \begin{bmatrix}
        M_{11}(\theta_a) & M_{12}(\theta_a) \\
        * & M_{22}(\theta_a)
    \end{bmatrix},
\end{equation*}
where
\begin{align*}
    \begin{split}
        M_{11}(\theta_a) &= m_1 R_1^2 + m_2 L^2 + m_3 \Big( L^2 \\
        &\quad + L R_3 \cos\Big(\frac{\theta_{a_1} - \theta_{a_2}}{2}\Big) + \frac{1}{4}R_3^2 \Big) \\
        &\quad + J_{zz_1} + \frac{1}{4} J_{zz_3},
    \end{split} \\
    \begin{split}
        M_{12}(\theta_a) &= m_2 L R_2 \cos(\theta_{a_1} - \theta_{a_2}) \\ 
        &\quad + m_3 \Big( L^2 \cos(\theta_{a_1} - \theta_{a_2}) \\
        &\quad + L R_3 \cos\Big(\frac{\theta_{a_1} - \theta_{a_2}}{2}\Big) + \frac{1}{4}R_3^2 \Big) \\
        &\quad + \frac{1}{4} J_{zz_3}, 
            \end{split} \\
    \begin{split}
        M_{22}(\theta_a) &= m_2 R_2^2 + m_3\Big( L^2 \\
        &\quad + L R_3 \cos\Big(\frac{\theta_{a_1} - \theta_{a_2}}{2}\Big) + \frac{1}{4}R_3^2 \Big) \\
        &\quad + J_{zz_2} + \frac{1}{4} J_{zz_3},
    \end{split}
\end{align*}
where $m_i$ with $i\in \{1,2,3\}$ denotes the mass of each link of the robot, $J_{zz_i}$ with $i\in \{1,2,3\}$ is the moment of inertia of each link with respect to the centers of mass of the links, $R_i$ with $i\in \{1,2,3\}$ is the distance to the center of mass of each link, $L$ denotes the length of each link, which is the same for both links and $\theta_a := [\theta_{a_1} \quad \theta_{a_1}]^\top \in \R^2$ is the link position of the robot. Finally, the nonlinear Coriolis and centrifugal matrix is given as 
\begin{equation*}
    \underline{C}(\theta_a,\dot{\theta}_a) :=
    \begin{bmatrix}
        C_1(\theta_a,\dot{\theta}_a) \\
        C_2(\theta_a,\dot{\theta}_a)
    \end{bmatrix},
\end{equation*}
where
\begin{align*}
    \begin{split}
        C_1(\theta_a,\dot{\theta}_a) &= -\frac{1}{4} m_3 L R_3 \sin\Big(\frac{\theta_{a_1} - \theta_{a_2}}{2}\Big)\dot{\theta}_{a_1}^2 \\
        &\quad + \Big(m_2 L R_2 \sin(\theta_{a_1} - \theta_{a_2}) \\ 
        &\quad + m_3 L^2 \sin(\theta_{a_1} - \theta_{a_2}) \\
        &\quad + \frac{3}{4} m_3 L R_3 \sin\Big(\frac{\theta_{a_1} - \theta_{a_2}}{2}\Big)\Big)\dot{\theta}_{a_2}^2 \\
        &\quad + \frac{1}{2} m_3 L R_3 \sin\Big(\frac{\theta_{a_1} - \theta_{a_2}}{2}\Big)\dot{\theta}_{a_1} \dot{\theta}_{a_2},
    \end{split}
\end{align*}
\begin{align*}
    \begin{split}
        C_2(\theta_a,\dot{\theta}_a) &= \frac{1}{4} m_3 L R_3 \sin\Big(\frac{\theta_{a_1} - \theta_{a_2}}{2}\Big)\dot{\theta}_{a_2}^2 \\
        &\quad - \Big(m_2 L R_2 \sin(\theta_{a_1} - \theta_{a_2}) \\ 
        &\quad + m_3 L^2 \sin(\theta_{a_1} - \theta_{a_2}) \\
        &\quad + \frac{3}{4} m_3 L R_3 \sin\Big(\frac{\theta_{a_1} - \theta_{a_2}}{2}\Big)\Big)\dot{\theta}_{a_1}^2 \\
        &\quad - \frac{1}{2} m_3 L R_3 \sin\Big(\frac{\theta_{a_1} - \theta_{a_2}}{2}\Big)\dot{\theta}_{a_1} \dot{\theta}_{a_2}.
    \end{split}
\end{align*}
The system parameters have been omitted due to confidentiality. 
\color{black}

\subsection{Power Transmission Dynamics} \label{app:gearbox_daynamics}
The power transmission dynamics as depicted in Figure~\ref{fig:harmonic_drive} can be derived using the Euler-Lagrange equation as given by
\begin{equation}
   \frac{d}{dt} ( \frac{\partial T}{\partial \dot{q}} )^\top - \frac{\partial T}{\partial q} + \frac{\partial V}{\partial q} = -\frac{\partial G}{\partial \dot{q}} + \tau_m,
   \label{eq:Lagrange}
\end{equation}
where $q = \theta_m$, $T$, and $V$ denote the kinetic and potential energy, respectively, and $G$ denotes the Rayleigh dissipation function, which are given by
\begin{align*}
   T &= \frac{1}{2} \dot{\theta}_m^\top \underline{J} \dot{\theta}_m, \\
   V &= \frac{1}{2} (\mu\theta_m - \theta_a)^\top \underline{K} (\mu\theta_m - \theta_a), \\
   G &= \frac{1}{2} (\mu \dot \theta_m - \dot{\theta}_a)^\top \underline{D} (\mu \dot \theta_m - \dot{\theta}_a).
\end{align*}
When now solving the Euler-Lagrange equation in \eqref{eq:Lagrange} the power transmission dynamics in \eqref{eq:harmonic_drive} can be obtained.

\subsection{Broken Belt Dynamics}\label{app:broken_belt}
The matrices in the faulty dynamics for the broken belt in \eqref{eq:faulty_dynamics_belt} are given as follows: 
\begin{equation*}
    \underline{M}_{f_1}(\theta_a, \theta_{a_3}) = 
        \begin{bmatrix}
            M_{f_{1_{11}}}(\theta_a) & M_{f_{1_{12}}}(\theta_a) & M_{f_{1_{13}}}(\theta_a, \theta_{a_3}) \\
            * & M_{f_{1_{22}}}(\theta_a) &M_{f_{1_{23}}}(\theta_a, \theta_{a_3}) \\
            * & * & M_{f_{1_{33}}}(\theta_a)
        \end{bmatrix}
\end{equation*}
with
\begin{equation*}
    \begin{aligned}
        M_{f_{1_{11}}}(\theta_a) &= m_1 R_1^2 + m_2 L^2 + m_3 L^2 + J_{zz_1},\\
        M_{f_{1_{12}}}(\theta_a) &= (m_2 R_2 + m_3 L) L \cos(\theta_{a_1}-\theta_{a_2}), \\
        M_{f_{1_{13}}}(\theta_a, \theta_{a_3}) &= m_3 L R_3 \cos(\theta_{a_1}-\theta_{a_3}),\\
        M_{f_{1_{22}}}(\theta_a) &= m_2 R_2^2 + m_3 L^2 + J_{zz_2},\\
        M_{f_{1_{23}}}(\theta_a, \theta_{a_3}) &= m_3 L R_3\cos(\theta_{a_2}-\theta_{a_3}),\\
        M_{f_{1_{33}}}(\theta_a) &= m_3 R_3^2 + J_{zz_3},
    \end{aligned}
\end{equation*}
\begin{equation*}
        \underline{C}_{f_1}(\theta_a,\dot{\theta}_a,\theta_{a_3},\dot{\theta}_{a_3}) = 
        \begin{bmatrix}
            C_{f_{1_1}}(\theta_a,\dot{\theta}_a,\theta_{a_3},\dot{\theta}_{a_3}) \\
            C_{f_{1_2}}(\theta_a,\dot{\theta}_a,\theta_{a_3},\dot{\theta}_{a_3}) \\
            C_{f_{1_3}}(\theta_a,\dot{\theta}_a,\theta_{a_3})
        \end{bmatrix}
\end{equation*}
with
\begin{equation*}
    \begin{aligned}
        C_{f_{1_1}}&(\theta_a,\dot{\theta}_a,\theta_{a_3},\dot{\theta}_{a_3}) = \\
        &m_{3}L^2\,{\dot{\theta}_{a_2}}^2\,\sin\left(\theta_{a_1}-\theta_{a_2}\right)\,+m_{2}\,R_{2}\,L\, {\dot{\theta}_{a_2}}^2 \sin\left(\theta_{a_1}-\theta_{a_2}\right)\,\\ 
        &+m_{3}\,R_{3}\,L\,{\dot{\theta}_{a_3}}^2\sin\left(\theta_{a_1}-\theta_{a_3}\right)\, \\
        C_{f_{1_2}}&(\theta_a,\dot{\theta}_a,\theta_{a_3},\dot{\theta}_{a_3}) = \\
        & -m_{3}\,L^2\,{\dot{\theta}_{a_1}}^2\sin\left(\theta_{a_1}-\theta_{a_2}\right)\,-m_{2}\,R_{2}\,L\,{\dot{\theta}_{a_1}}^2 \sin\left(\theta_{a_1}-\theta_{a_2}\right)\,\\
        &+m_{3}\,R_{3}\,L\,{\dot{\theta}_{a_3}}^2 \sin\left(\theta_{a_2}-\theta_{a_3}\right)\,,\\
        C_{f_{1_3}}&(\theta_a,\dot{\theta}_a,\theta_{a_3}) = \\
        &-m_{3}\,R_{3}\,L\,\left({\dot{\theta}_{a_1}}^2 \sin\left(\theta_{a_1}-\theta_{a_3}\right)\,+{\dot{\theta}_{a_2}}^2 \sin\left(\theta_{a_2}-\theta_{a_3}\right)\,\right), \\
    \end{aligned}
\end{equation*}
and 
\begin{equation*}
    \begin{aligned}
\underline{D}_{f_1}(\dot{\theta}_m, \dot{\theta}_a) &:=    \begin{bmatrix} \underline{D}(\dot{\theta}_a - \mu \dot{\theta}_m) \\ 0 \end{bmatrix}, \\
\underline{K}_{f_1}({\theta}_m, {\theta}_a) &:= \begin{bmatrix} \underline{K}({\theta}_a - \mu {\theta}_m) \\ 0 \end{bmatrix},\\ 
{\underline{D}_v}_{f_1} (\dot{\theta}_a) &:= \begin{bmatrix} \underline{D}_v \dot{\theta}_a \\ 0 \end{bmatrix}.
\end{aligned}
\end{equation*}

\subsection{Tilt Dynamics}\label{app:tilt}
The matrices in the faulty dynamics for the tilt scenario in \eqref{eq:faulty_dynamics_tilt} are given as follows: 
\begin{equation*}
    \underline{M}_{f_2}(\theta_a;\alpha,\beta,\gamma) = 
        \begin{bmatrix}
            M_{f_{2_{11}}}(\theta_a;\alpha,\beta,\gamma) & M_{f_{2_{12}}}(\theta_a;\alpha,\beta,\gamma) \\
            * & M_{f_{2_{22}}}(\theta_a;\alpha,\beta,\gamma) 
        \end{bmatrix}
\end{equation*}
with
\begin{equation*}
    \begin{aligned}
        M_{f_{2_{11}}}&(\theta_a;\alpha,\beta,\gamma) =  \\
        &J_{zz_1}+\frac{J_{zz_3}}{4}+(m_{3}+m_{2})\,L^2\,{\cos\left(\alpha \right)}^2\\
        &+m_{1}\,{R_{1}}^2\,{\cos\left(\alpha \right)}^2+\frac{m_{3}\,{R_{3}}^2\,{\cos\left(\gamma \right)}^2}{4}\\
        &+m_{3}\,L\,R_{3}\, \cos\left(\alpha \right)\,\cos\left(\gamma \right)\cos\big(\frac{\theta_{a_1} - \theta_{a_2}}{2}\big),\\
        M_{f_{2_{12}}}&(\theta_a;\alpha,\beta,\gamma) =  \\
        & \frac{J_{zz_3}}{4}+\frac{m_{3}\,{R_{3}}^2\,{\cos\left(\gamma \right)}^2}{4}\\
        &+m_{3}\,L^2\,\cos\left(\alpha \right)\,\cos\left(\beta \right)\,\cos\left(\theta_{a_1} - \theta_{a_2}\right)\,\\
        &+\frac{1}{2}{m_{3}LR_{3}(\cos\left(\alpha \right)+\cos\left(\beta \right))\cos\left(\gamma \right)\cos\big(\frac{\theta_{a_1} - \theta_{a_2}}{2}\big)}\\
        &+m_{2}\,L\,R_{2}\,cos\left(\alpha \right)\,\cos\left(\beta \right)\,\cos\big({\theta_{a_1} - \theta_{a_2}}\big),\\
        M_{f_{2_{22}}}&(\theta_a;\alpha,\beta,\gamma) = \\
        & J_{zz_2}+\frac{J_{zz_3}}{4}+m_{3}\,L^2\,{\cos\left(\beta \right)}^2+m_{2}\,{R_{2}}^2\,{\cos\left(\beta \right)}^2\\
        &+m_{3}\,\frac{{R_{3}}^2\,{\cos\left(\gamma \right)}^2}{4}\\
        &+m_{3}\,L\,R_{3}\,\cos\left(\beta \right)\,\cos\left(\gamma \right) \cos\big(\frac{\theta_{a_1} - \theta_{a_2}}{2}\big),\\
    \end{aligned}
\end{equation*}
and
\begin{equation*}
        \underline{C}_{f_2}(\theta_a,\dot{\theta}_a;\alpha,\beta,\gamma) =
        \begin{bmatrix}
            C_{f_{2_1}}(\theta_a,\dot{\theta}_a;\alpha,\beta,\gamma) \\
            C_{f_{2_2}}(\theta_a,\dot{\theta}_a;\alpha,\beta,\gamma) \\
        \end{bmatrix}
\end{equation*}
with
\begin{equation*}
    \begin{aligned}
        C_{f_{2_1}}&(\theta_a,\dot{\theta}_a;\alpha,\beta,\gamma) = \\
        -\frac{1}{4}&m_3 L R_3 \cos(\alpha) \cos(\gamma)\sin\Big(\frac{\theta_{a_1} - \theta_{a_2}}{2}\Big) \dot{\theta}_{a_1}^2 \\
        +\Big(&m_3 L^2 \sin \left({\theta_{a_1}}-{\theta_{a_2}}\right) \cos (\alpha) \cos (\beta) \\
        \quad + & m_2 L R_2 \sin \left({\theta_{a_1}}-{\theta_{a_2}}\right) \cos (\alpha) \cos (\beta) \\
        \quad + & (\frac{\cos(\alpha)}{4}+\frac{\cos(\beta)}{2}) m_3 L R_3 \cos(\gamma) \sin\left(\frac{\theta_{a_1} -\theta_{a_2} }{2}\right) \Big) \dot{\theta}_{a_2}^2 \\
        + &\frac{1}{2}{ m_3 L R_3\cos(\alpha) \cos(\gamma)} \sin\Big(\frac{\theta_{a_1} - \theta_{a_2}}{2}\Big) \dot{\theta}_{a_1} \dot{\theta}_{a_2},\\
        C_{f_{2_2}}&(\theta_a,\dot{\theta}_a;\alpha,\beta,\gamma) = \\
        +&\frac{1}{4}{m_3 L R_3  \cos(\beta) \cos(\gamma)} \sin\Big(\frac{\theta_{a_1} - \theta_{a_2}}{2}\Big)  \dot{\theta}_{a_2}^2\\
        -\Big( &m_3 L^2  \cos(\alpha) \cos(\beta) \sin\left({\theta_{a_1}} - {\theta_{a_2}}\right) \\
        \quad + &m_2 L R_2  \cos(\alpha) \cos(\beta) \sin\left({\theta_{a_1}} - {\theta_{a_2}}\right)\\
        \quad + &(\frac{\cos(\alpha)}{2}+\frac{\cos(\beta)}{4}) m_3 L R_3 \cos(\gamma) \sin\left(\frac{\theta_{a_1} -\theta_{a_2} }{2}\right) \Big) \dot{\theta}_{a_1}^2 \\
        -&\frac{1}{2}{m_3 L R_3  \cos(\beta) \cos(\gamma)} \sin\Big(\frac{\theta_{a_1} - \theta_{a_2}}{2}\Big) \dot{\theta}_{a_1} \dot{\theta}_{a_2}.
    \end{aligned}
\end{equation*}

\subsection{Proof of Theorem~\ref{theorem:optimal_estimator}} \label{ap:thm1_proof}

To prove this theorem, we provide three propositions, each of which guarantees one of the results of the theorem by providing an LMI-based condition. The proof of each proposition is provided in a separate appendix. Finally, using these propositions, the optimization problem in the theorem is formulated.
 
 Let us first introduce the following proposition, which formalizes an LMI-based condition that guarantees an ISS estimation error dynamics \eqref{eq:error_system} concerning input $(\omega_a, \nu_a)$. 
	\begin{prop}\emph{\textbf{(ISS Estimation Error Dynamics)} Consider the error dynamics \eqref{eq:error_system} and suppose there exist matrices $P \in {\mathbb{R}^{n_z \times n_z}}, P \succ0, R \in {\mathbb{R}^{n_z \times m}}$, and $ Q \in {\mathbb{R}^{n_z \times m}}$ satisfying the inequality
			\begin{equation}
				X 	 +  \epsilon I \preceq 0,
				\label{eq:stability_lmi}
			\end{equation}
			for some given $\epsilon>0$, with matrix $X$ as defined in \eqref{eq:X}, and $A_a$ and $C_a$ in \eqref{eq:augmented_matrices}. Then, the estimation error dynamics in \eqref{eq:error_system} is ISS with input $(\omega_a, \nu_a)$. Moreover, the ISS-gain from $(\omega_a, \nu_a)^\top$ to the estimation error $e$ in \eqref{eq:error_system} is upper bounded by $ 2 \|P {\left[\begin{array}{ccc} (I+E C_a) D_a & -K & E \end{array}\right]} \| \epsilon^{-1}$ with $E = P^{^{-1}} R $ and $K = P^{^{-1}} Q $.}
		\label{propos:stability}
	\end{prop}
	\emph{\textbf{Proof}}: 
The proof can be found in Appendix~\ref{ap:propos1_proof}.
\hfill $\blacksquare$

The second proposition formalizes an LMI-based condition guaranteeing $\|T_{e_d \omega_a}\|_{\infty}<\bar \lambda$, i.e., a finite $H_{\infty}$-norm of the transfer function $T_{e_d \omega_a}(s)$.
	
	\begin{prop}\emph{\textbf{(Finite $H_{\infty}$-norm)}
			Consider the estimation error dynamics \eqref{eq:error_system} with transfer matrix $T_{e_d \omega_a}(s)$ in \eqref{eq:tfs}. Assume there exist matrices $P\succ0, R,$ $Q$, and scalar $\bar \lambda > 0$ satisfying
			\begin{equation} \label{eq:hinf_lmi}
				\begin{aligned}
					&\left[\begin{array}{ccc}
						X & -(P + R C_a) D_a & \bar{C}_a^{\top} \\
						* & -\bar \lambda I  & {0} \\
						* & * & -\bar \lambda I
					\end{array}\right] \prec 0,\\
				\end{aligned}
			\end{equation}
			with $X$ defined in \eqref{eq:X}, $\bar{C}_a$ in \eqref{eq:c_bar_a}, and the remaining matrices in \eqref{eq:augmented_matrices}. Then, $\|T_{e_d \omega_a}\|_{\infty}<\bar \lambda$.}
		\label{propos:finite_Hinf}
	\end{prop}
		\emph{\textbf{Proof}}: The proof can be found in Appendix~\ref{ap:propos2_proof}.
	\hfill $\blacksquare$

The third proposition formalizes an LMI-based condition guaranteeing $\|T_{e_d \nu_a}\|_{H_2}<\bar \gamma$, i.e., a finite $H_2$-norm of $T_{e_d \nu_a}(s)$.
	\begin{prop}\emph{\textbf{(Finite $H_2$-norm)}
			Consider the estimation error dynamics \eqref{eq:error_system} with transfer matrix $T_{e_d \nu_a}(s)$ in \eqref{eq:tfs}. Assume there exist matrices $P\succ0, R,$ $Q$, $Z$, and scalar $\bar \gamma > 0$ satisfying
			\begin{equation} \label{eq:h2_lmi}
				\begin{aligned}
					& \left[\begin{array}{ccc}
						X & Q & -R \\
						* & -\bar \gamma I  & 0 \\
      				* & *  & -\bar \gamma I \\
					\end{array}\right] \prec 0, \quad\left[\begin{array}{cc}
						P & \bar{C}_a^{\top} \\
						* & Z
					\end{array}\right] \succ 0, \\
					&\operatorname{trace}(Z)<\bar \gamma,
				\end{aligned}
			\end{equation}
			with $X$ defined in \eqref{eq:X} and $\bar{C}_a$ in \eqref{eq:c_bar_a}. Then, \linebreak $\|T_{e_d \nu_a}\|_{H_2}<\bar \gamma$.}
		\label{propos:finite_H2}
	\end{prop}
		\emph{\textbf{Proof}}: The proof can be found in Appendix~\ref{ap:propos3_proof}. 
	\hfill $\blacksquare$

	Using Proposition~\ref{propos:finite_Hinf}, we can formulate a semi-definite program where we seek to minimize the $H_{\infty}$-norm of $T_{e_d \omega_a}(s)$. Similarly, using Proposition~\ref{propos:finite_H2}, we can formulate another semi-definite program where we seek to minimize the $H_2$-norm of $T_{e_d \nu_a}(s)$. However, in the presence of both unknown perturbations ($\omega_a$ and $\nu_{a}$), the $H_{\infty}$-norm and $H_2$-norm cannot be minimized simultaneously due to conflicting objectives. To attenuate the effect of measurement noise on the estimation errors, a relatively slow observer is required, which does not react to every small change in the measured variables. In contrast, to reduce the effect of $\omega_a$, a high-gain observer is preferred, which tries to estimate the fault as accurately as possible. It follows that there is a trade-off between estimation performance the noise sensitivity. To address this trade-off, a convex program where we seek to minimize the $H_{\infty}$-norm and constrain the $H_2$-norm,
	 for the same performance output $e_d$, is proposed. Moreover, we add the ISS LMI in \eqref{eq:stability_lmi} as a constraint to these programs to enforce that the resulting observer also guarantees boundedness for bounded perturbations and asymptotic stability for vanishing $\omega_a$ and $\nu_{a}$. The latter is essential to avoid observer divergence. The proof of Theorem~\ref{theorem:optimal_estimator} follows from the above analysis and Propositions~\ref{propos:stability}-\ref{propos:finite_H2}.

    \subsection{Proof of Proposition~\ref{propos:stability}} \label{ap:propos1_proof}
The following lemma is used to ensure ISS using an ISS Lyapunov function.
\begin{lem}\emph{\textbf{(ISS Lyapunov Function~{\cite[Thm. 4.19]{khalil2002nonlinear}})}
		Consider the error dynamics \eqref{eq:error_system} and let $W(e)$ be a continuously differentiable function such that
		\begin{equation*}
			\alpha_{1}(\|e\|) \leq W(e) \leq \alpha_{2}(\|e\|),
		\end{equation*}
		\begin{equation*}
			\dot{W}(e) \leq-W_{3}(e), \quad \hspace{-1mm} \forall \hspace{1mm}\|e\| \geq \chi\biggl(\biggl\| {\left[\begin{array}{cc}
					\omega_a(x, u, t)  \\ \nu_a(t)
				\end{array}\right]} \biggl\|\biggl),
		\end{equation*}
		where $\alpha_{1}(\cdot)$ and $\alpha_{2}(\cdot)$ are class $\mathcal{K}_{\infty}$ functions, $ \chi(\cdot)$ is a class $\mathcal{K}$ function, and $W_{3}$ is a continuous positive definite function. Then, the estimation error dynamics \eqref{eq:error_system} is ISS with gain $ {\eta}(\cdot) = \alpha_{1}^{-1}(\alpha_{2}({(\cdot)}))$.}
	\label{lem: iss}
\end{lem}

Let $W(e):={e}^\top P {e}$ be an ISS Lyapunov function candidate. Then, it follows from \eqref{eq:error_system} that
\begin{equation}
	\begin{aligned} \dot{W}(e) &\leq e^\top \Delta e  - 2 e^\top P
		\bar{B}_a
		{\left[\begin{array}{cc}
				\omega_a(x, u, t)  \\ \nu_a(t)
			\end{array}\right]},
	\end{aligned}
	\label{eq:lyapanov}
\end{equation}
where
\begin{equation}
	\begin{aligned} \Delta := &N^\top P+P N, \qquad
		\bar{B}_a := &{\left[\begin{array}{cc}
				M D_a & -\bar{B}
			\end{array}\right]}.
	\end{aligned}
	\label{eq:delta}
\end{equation}
Inequality \eqref{eq:lyapanov} implies the following 
\begin{equation}
	\begin{aligned} \dot{W}(e) \leq& -\lambda_{\min }(-\Delta) \|e\|^{2} + 2 \|e\| \|P \bar{B}_a\| \biggl\|{\left[\begin{array}{cc}
				\omega_a  \\ \nu_a
			\end{array}\right]}\biggl\| \\
		=& - (1-\theta) \lambda_{\min }(-\Delta) \|e\|^{2} - \theta \lambda_{\min }(-\Delta) \|e\|^{2} \\
		&+ 2 \|e\| \|P \bar{B}_a\|\biggl\|{\left[\begin{array}{cc}
				\omega_a  \\ \nu_a
			\end{array}\right]}\biggl\|,
	\end{aligned}
	\label{eq:iss_analysis}
\end{equation}
for any $\theta \in (0,1)$. Therefore, by \eqref{eq:iss_analysis} and Lemma~\ref{lem: iss}, if $\Delta$ is negative definite, error dynamics \eqref{eq:error_system} is ISS with input $(\omega_a, \nu_a)$ and linear ISS-gain
\begin{equation}
	\eta \biggl(\biggl\|{ \left[\begin{array}{cc}
			\omega_a  \\ \nu_a
		\end{array}\right]}\biggl \|\biggl)  = \frac{2 \|P \bar{B}_a\|}{\theta \lambda_{\min }(-\Delta)}\biggl\|{\left[\begin{array}{cc}
			\omega_a  \\ \nu_a
		\end{array}\right]}\biggl\|.
	\label{eq: gamma}
\end{equation}
Without loss of generality, for numerical tractability, we enforce $\Delta + \epsilon I \preceq 0$ for some arbitrarily small given $\epsilon >0$ instead of $\Delta \prec 0$. Now, we need to show that $\Delta$ is equivalent to $X$ in \eqref{eq:X} so that bt enforcing $X + \epsilon I \preceq 0$ in the proposition, the inequality above is satisfied.

Using $\Delta$ defined in \eqref{eq:delta} and \eqref{eq:observer_matrices}, we can write $\Delta$ in terms of the original observer gains $\{E,K\}$ as
\begin{equation*}
	\begin{aligned}
		\Delta = 	&A_{a}^\top(I+E C_{a})^\top P-C_{a}^\top  K P+P(I+E C_{a}) A_{a} \\
		&-P K C_{a}.
	\end{aligned}
\end{equation*}
Consider the following change of variables
\begin{equation}
	R :=P E,  \qquad	Q :=P K.
	\label{eq:variable_change}
\end{equation}
Applying \eqref{eq:variable_change} on the above expanded $\Delta$ leads to the same relation in \eqref{eq:X}. Therefore, we can conclude that $X$ is equivalent to $\Delta$. Clearly, $\Delta + \epsilon I \preceq 0$ implies $\lambda_{\min }(-\Delta) \geq \epsilon$. Then, using \eqref{eq: gamma}, we can conclude the bound on the ISS-gain in the statement of the proposition.

\subsection{Proof of Proposition~\ref{propos:finite_Hinf}}  \label{ap:propos2_proof}  
Let us first introduce the following lemma, in which we state a necessary and sufficient condition for having a bounded $H_{\infty}$-norm of $T_{e_d \omega_a}(s)$.
\begin{lem} \emph{\textbf{(Finite $H_{\infty}$-norm \cite[Prop. 3.12]{scherer2000linear})}
		Consider the estimation  error dynamics~\eqref{eq:error_system} with transfer matrix $T_{e_d \omega_a}(s)$ in \eqref{eq:tfs}. Assume $N$ is Hurwitz and consider a finite $\bar \lambda > 0$. Then, the following statements are equivalent:
		\begin{enumerate}
			\item $\|T_{e_d \omega_a}\|_{\infty}<\bar \lambda$.
			\item There exists a  $P \in {\mathbb{R}^{n_z \times n_z}}, P \succ 0$ satisfying
			\begin{equation} \label{eq:hinf_not_lmi}
				\left[\begin{array}{ccc}
					N^{\top} P+P N & -P M D_a & \bar{C}_a^{\top} \\
					* & -\bar \lambda I & {0} \\
					* & * & -\bar \lambda I
				\end{array}\right] \prec 0,
			\end{equation}
			with $\{N,M\}$ as in \eqref{eq:observer_matrices}, $\bar{C}_a$ as in \eqref{eq:c_bar_a}, and $D_a$ as in \eqref{eq:augmented_matrices}.
	\end{enumerate}}
	\label{lem:finite_Hinf}
\end{lem}
By applying the change of variables in \eqref{eq:variable_change}, one can see that \eqref{eq:hinf_lmi} is equivalent to \eqref{eq:hinf_not_lmi}, which, based on Lemma~\ref{lem:finite_Hinf}, allows to conclude the result of the proposition.

\subsection{Proof of Proposition~\ref{propos:finite_H2}}  \label{ap:propos3_proof}  
Let us first introduce the following lemma, in which we state a necessary and sufficient condition for having a bounded $H_2$-norm of $T_{e_d \nu_a}(s)$.
\begin{lem} \emph{\textbf{(Finite $H_2$-norm \cite[Prop. 3.13]{scherer2000linear})}
		Consider the estimation error dynamics \eqref{eq:error_system} with transfer matrix $T_{e_d \nu_a}(s)$ in \eqref{eq:tfs}. Assume $N$ is Hurwitz and consider a finite $\bar \gamma > 0$. Then, the following statements are equivalent:
		\begin{enumerate}
			\item $	\|T_{e_d \nu_a}\|_{H_2}<\bar \gamma$.
			\item There exists a  $P \succ 0$ and $Z$ satisfying
			\begin{equation} \label{eq:h2_not_lmi}
				\begin{aligned}
					& \left[\begin{array}{cc}
						N^{\top} P+P N & P \bar{B}\\
						* & -\bar \gamma I
					\end{array}\right] \prec 0, \\
     &\left[\begin{array}{cc}
						P & \bar{C}_a^{\top} \\
						* & Z
					\end{array}\right] \succ 0, \quad
					\operatorname{trace}(Z)<\bar \gamma,
				\end{aligned}
			\end{equation}
			with $N$ in \eqref{eq:observer_matrices}, $\bar{B}$ in \eqref{eq:b_bar}, and $\bar{C}_a$ in \eqref{eq:c_bar_a}.
	\end{enumerate}}
	\label{lem:finite_H2}
\end{lem}
By applying the change of variables in \eqref{eq:variable_change}, one can see that \eqref{eq:h2_lmi} is equivalent to \eqref{eq:h2_not_lmi}, which, based on Lemma~\ref{lem:finite_H2}, concludes the result of the proposition.

\subsection{Controller Structure}\label{ap:controller}
In the simulation environment a closed-loop feedback controller is used to ensure proper setpoint tracking of the model during simulation. To this end, proportional (P) and derivative (D) controller elements are employed to achieve sufficient setpoint tracking. The PD-controller can be characterized as
\begin{equation}
	u = 
	\begin{bmatrix}
		k_{p_1} & 0 \\
		0 & k_{p_2}
	\end{bmatrix}e_s +
	\begin{bmatrix}
		k_{d_1} & 0 \\
		0 & k_{d_2}
	\end{bmatrix}\dot{e}_s,
	\label{eq:PD-controller}
\end{equation}
where $u$ is the controller output, $e_s:=y-s$ is the setpoint tracking error, and $k_{p_s}$ and $k_{d_s}$ with $s \in \{1,2\}$ describe the proportional and derivative gains, respectively. The controller parameters have been omitted due to confidentiality.
\color{black}


 

\bibliographystyle{IEEEtran}
\bibliography{references}{}


 




\vfill

\end{document}